\documentclass{article}
\usepackage{ijcai21}

\usepackage{times}
\renewcommand*\ttdefault{txtt}
\usepackage{soul}
\usepackage{url}
\usepackage[hidelinks]{hyperref}
\usepackage[utf8]{inputenc}
\usepackage[small]{caption}
\usepackage{graphicx}
\usepackage{amsmath}
\usepackage{booktabs}
\usepackage{multirow}
\usepackage{hyperref}
\usepackage{url}
\usepackage{color}
\usepackage[utf8]{inputenc} 
\usepackage[T1]{fontenc}    
\usepackage{hyperref}       
\usepackage{url}            
\usepackage{booktabs}       
\usepackage{amsfonts}       
\usepackage{nicefrac}       
\usepackage{microtype}      

\usepackage{balance}
\usepackage{mathrsfs}
\usepackage{amsthm}
\usepackage{amsmath}
\usepackage{amssymb, bm}
\usepackage{multirow}
\usepackage{tabularx}
\usepackage{multicol}

\usepackage{graphicx}
\usepackage{subfig} 
\usepackage{changes}
\usepackage{wrapfig}
\newcommand{\yali}[1]{{ \color{magenta}[yali: #1]}}
\def\rc{\color{magenta}}

\usepackage{array}

\newtheorem{proposition}{Proposition}

\usepackage{algorithm}
\usepackage{algorithmic}

\title{Ordering-Based Causal Discovery with Reinforcement Learning}

\author{
	Xiaoqiang Wang$^{1}$\footnote{Work was done during an internship at Huawei Noah's Ark Lab.} \and
	Yali Du$^{2}$ \and
	Shengyu Zhu$^{3}$\footnote{Corresponding author.}\and\\ Liangjun Ke$^{1\dagger}$
	\and Zhitang Chen$^{3}$\and Jianye Hao$^{3,4}$
	\And
	Jun Wang$^2$\\
	\affiliations
	$^1$State Key Laboratory for Manufacturing Systems Engineering, School of Automation Science and Engineering, Xi'an Jiaotong University \\
	$^2$University College London \\
	$^3$Huawei Noah's Ark Lab \\ 
	$^4$College of Intelligence and Computing, Tianjin University
	\emails
	wangxq5127@{stu.}xjtu.edu.cn,
	yali.dux@gmail.com, zhushengyu@huawei.com,\\
	keljxjtu@xjtu.edu.cn, 
	\{chenzhitang2, haojianye\}@huawei.com, 
	jun.wang@cs.ucl.ac.uk
}

\begin{document}
	
	\maketitle
	
	\begin{abstract}
		It is a long-standing question to discover causal relations  among a set of variables in many empirical sciences. 
		Recently, Reinforcement Learning (RL)  has achieved promising results in causal discovery from observational data. However, searching the space of directed graphs and enforcing acyclicity by implicit penalties tend to be inefficient and restrict the existing RL-based method to small scale problems. 
		In this work, we propose a novel RL-based approach for causal discovery, by incorporating RL into the ordering-based paradigm.
		Specifically, we  formulate the ordering search problem as a multi-step Markov decision process, implement the ordering generating process with an encoder-decoder architecture, and finally use 
		RL to optimize the proposed model based on the reward mechanisms designed for~each ordering. 
		A generated ordering would then be processed using variable selection to obtain the final causal graph. 
		We  analyze the consistency and computational complexity of  the proposed method, and empirically show that a pretrained model can be exploited to accelerate training. Experimental results on both synthetic and real data sets shows  that the proposed method achieves a much improved performance over existing RL-based method.
	\end{abstract}
	
	\section{Introduction}
	
	Identifying causal structure from observational data is an important but also challenging task in many practical applications. This task can be formulated as that of finding a Directed Acyclic Graph (DAG) that minimizes a score function defined w.r.t.~the observed data. 
	However, 
	searching over the space of DAGs for the best DAG is known to be NP-hard, even if each node has at most two parents \cite{Chickering1996learning}.
	Consequently, traditional methods mostly rely on local heuristics to perform the search, including greedy hill-climbing and greedy equivalence search that explores the Markov equivalence classes \cite{chickering2002optimal}. 
	
	Along with various search strategies, existing methods have also cast causal structure learning problem as that of learning an
	optimal variable ordering,  considering that the ordering space is significantly smaller than that of directed graphs and searching over the ordering space can avoid dealing with the acyclicity constraint \cite{teyssier2012ordering}.
	%
	%
	Many methods, such as genetic algorithm \cite{larranaga1996learning}, Markov chain Monte Carlo \cite{friedman2003being} and greedy local hill-climbing \cite{teyssier2012ordering}, have been exploited as the search strategies to find desired orderings. In practice, however, 
	these methods often cannot effectively find a globally optimal  ordering   for their heuristic nature. 
	
	Recently, with smooth score functions, several gradient-based methods have been proposed by exploiting a smooth characterization of acyclicity, including NOTEARS \cite{zheng2018dags} for linear causal models and several subsequent works, e.g., \cite{yu19dag,Lachapelle2019grandag,Ng2019GAE,Ng2019masked,Zheng2019learning}, which use neural networks for modelling non-linear causal relationships. As another attempt,  \cite{zhu2020causal} utilize Reinforcement Learning (RL) to find the underlying DAG from the graph space without the need of smooth score functions.  
	Unfortunately, \cite{zhu2020causal} achieved good performance only with up to $30$ variables, for at least two reasons:~1) the action space, consisting of directed graphs, is tremendous for large scale problems and is hard to be explored efficiently; and 2) it has to compute scores for many non-DAGs generated during training but computing scores w.r.t.~data is generally time-consuming.
	It appears that the RL-based approach may not be able to achieve a close performance to other gradient-based methods that directly optimize the same 
	score function for large causal discovery problems,  due to its search nature.


	By taking advantage of the reduced space of variable orderings and the strong search ability of modern RL methods, we propose Causal discovery with Ordering-based Reinforcement Learning (CORL), which incorporates RL into the ordering-based paradigm and is shown to achieve a promising empirical performance. 
	In particular, CORL outperforms NOTEARS, a state-of-the-art gradient-based method for linear data, even with $150$-node graphs. Meanwhile, CORL is also competitive with a strong baseline, Causal Additive Model (CAM) method \cite{buhlmann2014cam}, on non-linear data models.
	
	\paragraph{Contributions.} We make the following contributions in this work: 1) We  formulate the ordering search problem as a multi-step Markov Decision Process (MDP) and  propose to implement the ordering generating process in an effective encoder-decoder architecture, followed by applying RL to optimizing the proposed model based on specifically designed reward mechanisms. We also incorporate a pretrained model into CORL to accelerate training. 
	%
	%
	%
	2) We analyze the consistency and computational complexity  of the proposed method. 
	3) We conduct comparative experiments on synthetic and real data sets to validate the performance of the proposed methods. 4) An  implementation  has been made  available at \url{https://github.com/huawei-noah/trustworthyAI/tree/master/gcastle}.\footnote{The extended version can be found at \url{ https://arxiv.org/abs/2105.06631.}}

	\section{Related Works}
	\label{relatedworks}
	Besides the aforementioned heuristic ordering search algorithms, \cite{schmidt2007learning} proposed L1OBS to conduct variable selection using $\ell_1$-regularization paths based on the method from \cite{teyssier2012ordering}. 
	\cite{scanagatta2015learning} further proposed an ordering exploration method on the basis of an approximated score function so as to scale to thousands of variables. 
	The CAM method \cite{buhlmann2014cam} was specifically designed for non-linear additive models. Some recent ordering-based methods such as sparsest  permutation  \cite{raskutti2018learning} and  greedy {sparsest} permutation  \cite{solus2017consistency} can guarantee consistency of Markov equivalence class, relying on some conditional independence relations and  certain assumptions like faithfulness. A variant of  greedy sparest permutation was further proposed in \cite{bernstein2020ordering} for the setting with latent variables. 
	In the present work, we mainly work on identifiable cases which have different assumptions from theirs.
	
	
	In addition, exact algorithms such as dynamic programming \cite{Xiang2013lasso} and integer or linear programming \cite{bartlett2017integer} are also used for causal discovery problem. 
	However, these algorithms usually can only work on small graphs \cite{de2011efficient}, and to handle larger problems with hundreds of variables, they usually need to incorporate heuristics search \cite{Xiang2013lasso} or limit the maximum number of parents of each node.

	Recently, RL has been used to tackle several combinatorial problems such as the maximum cut and traveling salesman problem \cite{bello2016neural,khalil2017learning,kool2018attention}.
	These works aim to learn a policy as a solver based on {the particular type of combinatorial problems}. 
	However,  causal discovery tasks generally have different relationships, data types, graph structures, etc., and moreover, are typically off-line with focus on a or a class of causal graph(s). As such, we use RL as a search strategy, similar to \cite{zoph2016neural,zhu2020causal}. Nevertheless, a pretrained model or policy can offer a good starting point to speed up training, as shown in our evaluation results (cf.~Figure~\ref{reward-curve-corl2}). 
	

	\section{Background}
	\label{gen_inst}
	\subsection{Causal Structure Learning}
	
	Let $\mathcal{G}=(d, V, E)$ denotes a DAG, with $d$ the number of nodes, $V=\{v_1,\cdots,v_d\}$ the set of nodes, and $E=\{(v_i,v_j) | i,j= 1,\ldots,d\}$ the set of directed edges from $v_i$ to $v_j$. Each node $v_j$ is associated with a random variable $X_j$. 
	The probability model associated with  $\mathcal{G}$ factorizes as $p(X_1,\cdots,X_d)=\prod^d_{j=1}p(X_j|\text{Pa}(X_j))$, where $p(X_j|\text{Pa}(X_j))$ is the conditional probability distribution for $X_j$ given
	its parents $\text{Pa}(X_j):=\{X_k|(v_k, v_j)\in E\}$. 
	We assume that the observed data $\mathbf {x}_j$ is obtained
	by the Structural Equation Model (SEM) with {additive} noises:  $X_j := f_j(\text{Pa}(X_j)) + \epsilon_j, j=1,\ldots,d$, where $f_j$ represents the functional relationship between $X_j$ and its parents, and $\epsilon_j$'s denote jointly independent additive noise variables. 
	We assume causal minimality, which is equivalent to that each $f_j$ is not a constant for any $X_k \in \text{Pa}(X_j)$ in this SEM \cite{10.5555/2627435.2670315}.
	
	Given a sample $\mathbf{X} = [\mathbf {x}_1,\cdots,\mathbf{x}_d] \in\mathbb R^{m\times d}$ where $\mathbf{x}_j$ is a vector of $m$ observations for random variable $X_j$. The goal is to find a DAG  $\mathcal{G}$ that optimizes the  Bayesian Information Criterion (BIC) (or equivalently, minimum description length) score, defined as 
	\begin{equation}\label{eq4-0}
	\text{S}_\text{BIC}\mathcal{(G)}\hspace{-2pt}=\hspace{-2pt}\sum^d_{j=1}\hspace{-2pt}\left[\sum^m_{k=1} \log p(x^k_j|\text{Pa}(x^k_j);\theta_j)\hspace{-2pt}-\hspace{-2pt}\frac{|\theta_j|}{2} \log m \right]\hspace{-2pt},
	\end{equation}
	where $x^k_j$ is the $k$-th observation of $X_j$, $\theta_j$ is the parameter associated with each likelihood, and $|\theta_j|$ denotes the parameter dimension. For linear-Gaussian models, 
	$p(x^k_j|\text{Pa}(x^k_j);\theta_j)= \mathcal{N}(x_j|\theta_j^T\text{Pa}(x_j), \sigma^2)$ and $\sigma^2$ can be estimated from the data. 
	
	\begin{figure}
		\centering
		\includegraphics[width=0.15 \textwidth]{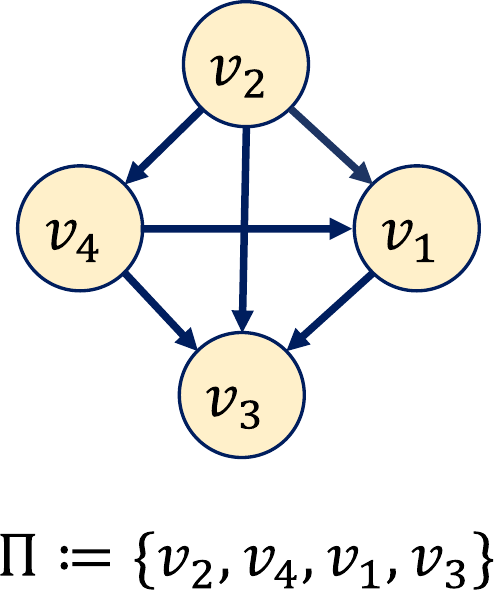}
		\caption{An example of the correspondence between an ordering  and a fully-connected DAG.}
		\label{model_dag}
	\end{figure}
	
	The problem of finding a directed graph that satisfies the ayclicity constraint
	can be cast as that of finding a  variable ordering \cite{teyssier2012ordering,schmidt2007learning}.  
	Specifically, let $\Pi$ denote an ordering of the nodes in $V$, where the length of the ordering $|\Pi|=|V|$ and $\Pi$ is indexed from 1. If node $v_j\in V$ lies in the $p$-th position, then $\Pi(p)=v_j$. Notation $\Pi_{\prec v_j}$ denotes the set of nodes  that precede node $v_j$ in $\Pi$. One can easily establish a canonical correspondence between an ordering $\Pi$ and a fully-connected DAG $\mathcal{G}^\Pi$; an example  is presented in Figure~1. 
	A DAG $\mathcal{G}$ can be consistent with more than one orderings and the set of these orderings is denoted by
	\begin{equation*}
	\Phi(\Pi)\hspace{-2pt}=\hspace{-2pt}\{\Pi:~\text{fully-connected DAG}~\mathcal{G}^{\Pi}~\text{is a super-DAG of} \ \mathcal{G} \},
	\end{equation*}
	where a super-DAG of $\mathcal{G}$ is a DAG whose edge set is a superset of that of $\mathcal{G}$.  
	The the search for the true DAG $\mathcal{G}^{*}$ can be decomposed to two phases: finding the correct ordering and performing variable selection; the latter is to find the optimal DAG that is consistent with the ordering found in the first step.

	
	\subsection{Reinforcement Learning}
	Standard RL is usually formulated as an MDP over the environment state $s \in \mathcal{S}$ and agent action
	$a\in \mathcal{A}$, under an (unknown) environmental dynamics defined by a transition probability $\mathcal{T}(s'|s,a)$. Let $\pi_{\phi}(a|s)$ denote the policy,  parameterized by $\phi$, which outputs a  distribution used to select an action from action space $\mathcal{A}$ based on state $s$. For episodic tasks, a trajectory $\tau = \{s_t,a_t\} ^T_ {t=0}$, where $T$ is the finite time horizon, can be collected by executing the policy repeatedly. In many cases, an immediate reward $r(s,a)$ can be received when agent executes an action.
	The objective of RL is  to learn a policy which can maximize the expected cumulative reward along a trajectory, i.e., $J(\phi) =\mathbb{E}_{\pi_{\phi}}[R_0]$ with $R_0=\sum^{T}_{t=0}\gamma^{t}r_{t}(s_{t}, a_{t})$ and $\gamma \in (0,1]$ being a discount factor. For some scenarios, the reward is only earned at the terminal time (also called episodic reward), 
	and $J(\phi) =\mathbb{E}_{\pi_{\phi}}[R(\tau)]$ with $R(\tau)=r_{T}(s_T, a_T)$.

	
	\section{Method}
	In this section, we first formulate the ordering search problem as an MDP and then describe the proposed approach. We also discuss the variable selection methods to obtain DAGs from variable orderings, as well as the consistency and computational complexity regarding the proposed method.
	
	\subsection{Ordering Search as Markov Decision Process}
	To incorporate RL into the ordering-based paradigm, we formulate the variable ordering search problem as a multi-step decision process with a variable as  an action at each decision step, and the order of the selected actions (or variables) is treated as the searched ordering. The decision-making process is Markovian, and its elements are described as follows.
	
	\paragraph{State.}
	One can directly take the sample data $\mathbf{x}_j$ as the state. However, preliminary experiments (see Appendix~A.1) show that it is difficult for feed-forward neural network models to capture the underlying causal relationships directly using observed data as states,  
	and that the data pre-processed by an encoder module is helpful to find better orderings. 
	The encoder module embeds each $\mathbf{x}_j$ to state $s_j$ and all the embedded states constitute the state space $\mathcal{S}:=\{ {s}_1,\cdots, {s}_d\}$. In our case, we also need an initial state, denoted by $s_0$ (detailed choice is given in Section~4.2),  to select the first action. The complete state space would be $\mathcal{\hat{S}}:= \mathcal{S}\cup \{s_0\}$. We will use $\hat{s}_t$ to denote the actual state encountered at the $t$-th decision step when generating a variable ordering. 

	
	\paragraph{Action.}
	We select an action (variable) from the action space {$\mathcal{A}:=\{ {v}_1,\cdots, {v}_d\}$} consisting of all the variables at each decision step, and the action space size is equal to the number of variables, i.e., $|\mathcal{A}|=d$. Compared to the previous RL-based method that searches over the graph space with size $\mathcal{O}(2^{d\times d})$ \cite{zhu2020causal}, the resulting action space becomes much smaller. 
	
	\paragraph{State transition.}
	The state transition is related to the action selected at the current decision step. If the selected variable is $v_j$ at the $t$-th decision step, 
	then the state is transferred to the state $s_j \in \mathcal{S}$ which 
	corresponds to $\mathbf{x}_j$ embedded by the encoder, 
	i.e., $\hat{s}_{t+1}=s_j$.
	
	\paragraph{Reward.}
	In ordering-based methods, only the variables selected in previous decision steps can be the potential parents of the currently selected variable. Hence, we  design the rewards in the following cases: \textit{episodic reward} and \textit{dense reward}. 
	In the former case, we   calculate the score for a  variable ordering $\Pi$ {with $d$ variables} as the episodic reward, i.e., 
	\begin{equation}
	R(\tau)=r_{T}(\hat{s}_T, a_T)= \text{S}_\text{BIC}(\mathcal{G}^{\Pi})
	\end{equation}
	where $T = d-1$ and $\text{S}_\text{BIC}$ has been defined in Equation~(1), with $\text{Pa}(X_j)$  replaced by the potential parent variable set $U(X_j)$; here $U(X_j)$ denotes the variables associated with the nodes in $\Pi_{\prec v_j}$. 
	If the score function is decomposable (e.g., the BIC score), 
	we can calculate an immediate reward by exploiting the decomposability for the current decision step. That is, for $v_j$ selected at time step $t$, the immediate reward is 
	\begin{equation}
	r_t = \sum^m_{k=1} \log p(x^k_j|U(x_j^{k});\theta_j)  - \frac{|\theta_j|}{2} \log m . 
	\end{equation}
	This belongs the second case with {\it dense rewards}. {Here we keep  $-|\theta_j|/2 \log m$  to make  Equation~(3) consistent with the form of BIC score. }
	
	\subsection{Implementation and Optimization with Reinforcement Learning}
	\begin{figure}
		\centering
		\includegraphics[width=0.33\textwidth,]{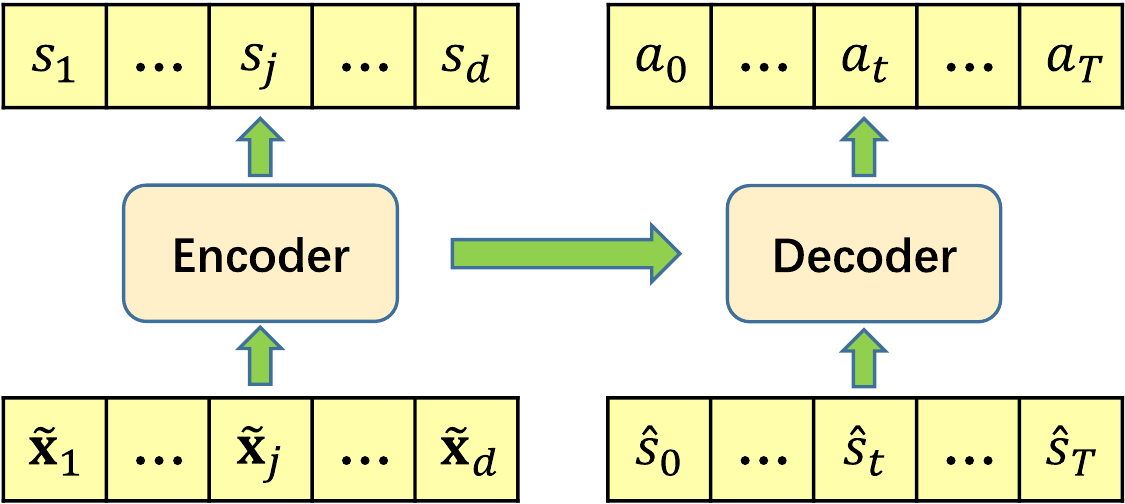}
		\caption{Illustration of the policy model. The encoder embeds the observed data $\mathbf{x}_j$ into the state $s_j$. An action $a_t$ can be selected by the decoder according to the given state $\hat{s}_t$ and the pointer mechanism at each time step $t$. Note that $T = d-1$. See Appendix A for details.} \label{model_stru}
	\end{figure}
	We briefly describe the neural network architectures implemented in our method, as shown in Figure~2.
	More details can be found in Appendix~A.
	
	
	\paragraph{Encoder.} 
	$f_{\phi_e}^{\mathrm{enc}}: \tilde{\mathbf{X}}  \mapsto \mathcal{S}$  is used to map the observed data to the embedding space $\mathcal{S}=\{s_1,\cdots, s_d\}$. Similar to \cite{zhu2020causal}, we adopt mini-batch training and randomly draw $n$ samples from $m$ samples of the data set $\mathbf{X}$  to construct $\tilde{\mathbf{X}}\in \mathbb R^{n\times d}$ at each episode. 
	We also set the embedding $s_j$ to be in the same dimension, i.e., $s_j\in\mathbb R^n$. For encoder choice, we conduct an empirical comparison among several representative structures such as MLP, LSTM and the self-attention based encoder \cite{vaswani2017attention}. Empirically, we validate that the self-attention based encoder in the Transformer structure performs the best (see Appendix~A.1).
	
	\paragraph{Decoder.} $f_{\phi_d}^{\text{dec}}: \mathcal{\hat{S}} \mapsto \mathcal{A}$ maps the state space $\mathcal{\hat{S}}$ to the action space $\mathcal{A}$. Among several decoder choices (see also Appendix~A.1 for an empirical comparison), we pick an LSTM based structure that proves effective in our experiments. 
	Although the initial state is  generated randomly in many applications, we  pick it as $s_0=\frac1d\sum_{i=1}^d s_i$, considering that the source node is fixed in a correct ordering.  
	We  restrict each node only be selected once by masking the selected nodes,  in order to generate a valid ordering \cite{vinyals2015pointer}. 
	
	\paragraph{Optimization.} The optimization objective is to learn a policy maximizing
	$J(\phi)$,
	where $\phi=\{\phi_e,\phi_d\}$ with $\phi_e$ and $\phi_d$ being parameters associated with encoder $f^{\text{enc}}$ and decoder $f^{\text{dec}}$, respectively.
	Based on the above definition, policy gradient \cite{sutton2018reinforcement} is used to optimize the ordering generation model parameters. 
	For the \textit{episodic reward} case, we have the following policy gradient $\nabla J(\phi)=\mathbb{E}_{\pi_{\phi}}\left[R(\tau) \sum^T_{t=0} \nabla_{\phi} \log \pi_{\phi}\left(a_t| \hat{s}_t\right) \right]$, and the algorithm in this case is denoted as CORL-1. 
	For the \textit{dense reward} case, policy gradient can be calculated as $\nabla J(\phi)=\mathbb{E}_{\pi_{\phi}}\left[\sum^T_{t=0}R_t \nabla_{\phi} \log \pi_{\phi}\left(a_t| \hat{s}_t\right) \right]$,
	where $R_t=\sum^{T-t}_{l=0}\gamma^{l}r_{t+l}$ denotes the return at time step $t$.  We denote the algorithm in this case as CORL-2. 
	Using a parametric baseline to estimate the expected score typically improves
	learning \cite{sutton2018reinforcement}. Therefore, we introduce a critic network $V_{\phi_{v}}(\hat{s}_t)$ parameterized by $\phi_{v}$, which learns the expected return given state $\hat{s}_t$ and is trained with stochastic gradient descent using Adam optimizer on a mean squared error objective between its
	predicted value and the actual return. More details about the critic network are described in  Appendix~A.2.
	
	Inspired by the benefits from pretrained models \cite{bello2016neural}, we also consider to incorporate pretraining to our method to accelerate training. In practice, one can usually obtain some observed data with known causal graphs or  correct orderings, e.g., by simulation or real data with labeled graphs. Hence, we can pretrain a policy model with such data in a supervised way and use the pretrained model as initialization for new tasks. Meanwhile, a sufficient generalization ability is desired and we hence include diverse data sets with different numbers of nodes, noise types, causal relationships, etc.
	
	
	\subsection{Variable Selection}
	\begin{table*}[ht]
		\linespread{1.0} 
		\centering \scriptsize 
		\begin{tabular}{lcccccccccc}
			\toprule 
			&  & & RANDOM & NOTEARS & DAG-GNN & RL-BIC2  & L1OBS & A{*} Lasso & CORL-1 & CORL-2 \\
			\midrule
			\multirow{8}{*}{30 nodes} &\multirow{2}{*}{ER2} & TPR & 0.41 (0.04) & 0.95 (0.03) & 0.91 (0.05) & 0.94 (0.05) &  0.78 (0.06) & 0.88 (0.04) & \bf{0.99 (0.02)}&\bf{0.99 (0.01)} \\
			& & SHD& 140.4 (36.7) & 14.2 (9.4) & 26.5 (12.4)  & 17.8 (22.5) & 85.2 (23.8) &35.3 (14.3) & \bf{5.2 (7.4)} & \bf{4.4 (3.5)}\\
			\cmidrule(){2-11}  
			& \multirow{2}{*}{ER5} & TPR & 0.43 (0.03) & \bf{0.93 (0.01)} & 0.85 (0.11)  & 0.91 (0.03)  & 0.74 (0.04)  &  0.84 (0.05) & \bf{0.94 (0.03)} & \bf{0.95 (0.03)} \\
			&  & SHD & 210.2 (43.5) &  \bf{35.4 (7.3)} &  68.0 (39.8) & {45.6 (13.3)}  &  98.6 (32.7) &  71.2 (21.5) & \bf{37.4 (16.9)} & \bf{37.6 (14.5)}\\
			\cmidrule(){2-11}
			& \multirow{2}{*}{SF2} & TPR & 0.58 (0.02) & 0.98 (0.02) & 0.92 (0.09)  & 0.99 (0.02)  & 0.83 (0.04)  & 0.93 (0.02)  & \bf{1.0 (0.01)} & \bf{1.0 (0.01)} \\
			&& SHD & 118.4 (12.3) & 6.1 (2.3) & 36.8 (33.1)  & 3.2 (1.7)  & 49.7 (28.1)  & 27.3 (18.4)  & \bf{0.0 (0.0)} & \bf{0.0 (0.0)}\\
			\cmidrule(){2-11}
			& \multirow{2}{*}{SF5} & TPR & 0.44 (0.03) & 0.94  (0.03) & 0.89 (0.09)  & 0.96  (0.03)  & 0.79 (0.04)  & 0.88 (0.03)  & \bf{1.00 (0.00)} & \bf{1.00 (0.00)} \\
			&& SHD & 165.4 (10.6) & 23.3 (6.9) & 47.8 (35.2)  & 11.3 (5.2)  & 89.3 (25.7)   & 40.5 (19,8)  & \bf{0.0 (0.0)} & \bf{0.0 (0.0)}\\
			
			\midrule
			\multirow{8}{*}{100 nodes}  &\multirow{2}{*}{ER2} & TPR & 0.33 (0.05) & 0.93 (0.02) & 0.93 (0.03) & 0.02 (0.01) &  0.54 (0.02) & 0.86 (0.04) & \bf{0.98 (0.02)} & \bf{0.98 (0.01)} \\
			&  & SHD  & 491.4 (17.6) & 72.6 (23.5) & 66.2 (19.2) & 270.8 (13.5) & 481.2 (49.9) & 128.5 (38.4) & \bf{24.8 (10.1)}& \bf{18.6 (5.7)}\\
			\cmidrule(){2-11}
			&  \multirow{2}{*}{ER5} & TPR & 0.34 (0.04) & \bf{0.91 (0.01)}& 0.86 (0.16) & 0.08 (0.03) & {0.53 (0.02)} & {0.82 (0.05)} & \bf{0.93 (0.02)} & \bf{0.94 (0.03)} \\
			&  & SHD & 984.4 (35.7) & \bf{170.3 (34.2)} & {236.4 (36.8)} & {421.2 (46.2)}   & {547.9 (63.4)} & {244.0 (42.3)} &\bf{175.3 (18.9)} & \bf{164.8 (17.1)}\\
			\cmidrule(){2-11}
			& \multirow{2}{*}{SF2} & TPR & 0.48 (0.03)  & 0.98 (0.01) & 0.89 (0.14) & {0.04 (0.02)} & {0.57 (0.03)} & {0.92 (0.03)} & \bf{1.00 (0.00)} & \bf{1.00 (0.00)} \\
			&& SHD & 503.4 (23.8) & 2.3 (1.3) & {156.8 (21.2)} & {281.2 (17.4)}   & {377.3 (53.4)} & {54.0 (22.3)} &\bf{0.0 (0.0)} & \bf{0.0 (0.0)}\\
			\cmidrule(){2-11}
			& \multirow{2}{*}{SF5} & TPR & 0.47 (0.04) & 0.95 (0.01) &{0.87 (0.15)} & {0.05 (0.03)} & {0.55 (0.04)} & {0.89 (0.03)} & \bf{0.97 (0.02)} & \bf{0.98 (0.01)} \\
			&& SHD & 891.3 (19.4) & 90.2 (34.5) & 165.2 (22.0) & {405.2 (77.4)}   & {503.7 (56.4)} & {114.0 (36.4)} &\bf{19.4 (5.2)} & \bf{10.8 (6.1)}\\
			\bottomrule
		\end{tabular}
		\caption{Empirical results for ER and SF graphs of $30$ and $100$ nodes with LG data.}\label{30-Lin_results}
	\end{table*} 
	
	One can  obtain the causal graph from an ordering by conducting variable selection methods, such as sparse candidate \cite{teyssier2012ordering},  significance testing of covariates \cite{buhlmann2014cam}, and  group Lasso \cite{schmidt2007learning}. 
	In this work, 
	for linear data models, we apply linear regression to the obtained fully-connected DAG and then use thresholding to prune edges with small weights, as similarly used by \cite{zheng2018dags}.
	For the non-linear model, we adopt  the CAM pruning used by \cite{Lachapelle2019grandag}. For each variable $X_j$, one can fit a generalized additive model against the
	current parents of $X_j$ and then apply significance testing of covariates, 
	declaring significance if the reported p-values are lower that or equal to $0.001$. The overall method is summarized in Algorithm 1. 
	
	\begin{algorithm}[t]
		\label{alg:all} 
		\caption{{Causal discovery with Ordering-based RL.} 
		}
		\begin{algorithmic}[1]
			\REQUIRE observed data $\mathbf{X}$, initial parameters $\phi_e,\phi_d$ and $\phi_v$, two empty buffers $\mathcal{D}$ and $\mathcal{D}_{score}$, initial value (negative infinite) BestScore and a random ordering  BestOrdeing.
			\WHILE{not terminated}
			\STATE draw a batch of samples from $\mathbf{X}$,   		encode them to $\mathcal{S}$ and calculate the initial state $\hat{s}_0$
			\FOR{$t=0,1,\ldots,T$}
			\STATE collect a batch of data $\langle \hat{s}_t,a_t,r_t\rangle$ with $\pi_{\phi}$: $\mathcal{D} = \mathcal{D} \cup \{\langle \hat{s}_t,a_t,r_t\rangle\}$
			\IF{$\langle v_t,\Pi_{\prec v_t}, r_t \rangle$ is not in $\mathcal{D}_{score}$}
			\STATE store $\langle v_t,\Pi_{\prec v_t}, r_t \rangle$  in $\mathcal{D}_{score}$ 
			\ENDIF
			\ENDFOR
			\STATE {update $\phi_e,\phi_d$, and $\phi_v$ as described in Section~4.2
			} \IF{$\sum^T_{t=0}r_t > \text{BestScore}$}
			\STATE update the BestScore and BestOrdering
			\ENDIF
			\ENDWHILE
			\STATE get the final DAG by pruning the BestOrdering
		\end{algorithmic}
	\end{algorithm}
	\subsection{Consistency Analysis}
	So far we have presented CORL in a general manner without specifying explicitly
	the distribution family for calculating the scores or rewards. In principle, any distribution family could be employed
	as long as its log-likelihood can be computed. 
	However, whether the maximization of the  accumulated reward recovers the correct ordering, i.e., whether  consistency of the score function holds, depends on both the modelling choice of reward and the underlying SEM. If the SEM is identifiable, then the following proposition shows that it is possible to find the correct ordering with high probability in the large sample limit.
	\begin{proposition}
		Suppose that an identifiable SEM with true causal DAG $\mathcal{G}^{*}$ on $X=\{X_j\}_{j=1}^d$ induces distribution $P(X)$. Let $\mathcal{G}^{\Pi}$ be the fully-connected DAG that corresponds to an ordering $\Pi$. If there is an SEM with $\mathcal{G}^{\Pi}$ inducing the same distribution $P(X)$, then $\mathcal{G}^{\Pi}$ must be a super-graph of $\mathcal{G}^{*}$, i.e., every edge in $\mathcal{G}^{*}$ is covered in $\mathcal{G}^{\Pi} $.
	\end{proposition}
	\begin{proof}
		The SEM with $\mathcal{G}^{\Pi}$ may not be causally minimal but can be reduced to an SEM satisfying the causal minimality condition \cite{10.5555/2627435.2670315}. Let $\tilde{\mathcal{G}}^{\Pi}$ denotes the causal graph in the reduced SEM with the same distribution $P(X)$. Since we have assumed that original SEM is identifiable, i.e., the distribution $P(X)$ corresponds to a unique true graph, $\tilde{\mathcal{G}}^{\Pi}$ is then identical to $\mathcal{G}^{*}$. The proof is complete by noticing that $\mathcal{G}^{\Pi}$ is a super-graph of $\tilde{\mathcal{G}}^{\Pi}$. 
	\end{proof}
	Thus, if the causal relationships fall into the chosen model functions and a right distribution family is assumed, then given infinite samples the optimal accumulated reward (e.g., the optimal BIC score) must be achieved by a super-DAG of the underlying graph.  
	However, 
	finding the optimal accumulated reward  may be hard,
	because policy gradient methods only guarantee local convergence  \cite{sutton2018reinforcement}, and  we can only apply approximate model functions and also need to assume a certain distribution family for calculating the reward.  
	Nevertheless, the  experimental results in Section~5 show that the proposed method can achieve a better performance than those with consistency guarantee in the finite sample regime, thanks to the improved search ability of modern RL methods.

	\subsection{Computational Complexity}

	In contrast with typical RL applications, we treat RL here as a search strategy
	, aiming to find an ordering that achieves the best score. 
	CORL requires the evaluation of the rewards at each episode with $\mathcal{O}(dm^2+d^{3})$ computational cost if linear functions are adopted to model the causal relations, which is same to RL-BIC2 \cite{zhu2020causal}. Fortunately, CORL does not need to compute the matrix exponential term with $\mathcal O(d^3)$  cost due to the use of ordering search.
	We observe that CORL performs fewer episodes than RL-BIC2 before the episode reward converges (see Appendix~C). 
	The evaluation of Transformer encoder and LSTM decoder in CORL take $\mathcal{O}(nd^{2})$ and $\mathcal{O}(dn^{2})$, 
	respectively. However, we find that computing rewards is dominating in the total running time (e.g., around $95\%$ and $87\%$ for $30$- and $100$-node linear data models). Thus, we record the decomposed scores for each variable $v_j$ with different parental sets $\Pi_{\prec v_j}$ to avoid repeated computations. 

	
	\section{Experiments}
	\begin{figure}
		\centering
		\includegraphics[width=0.46\textwidth,]{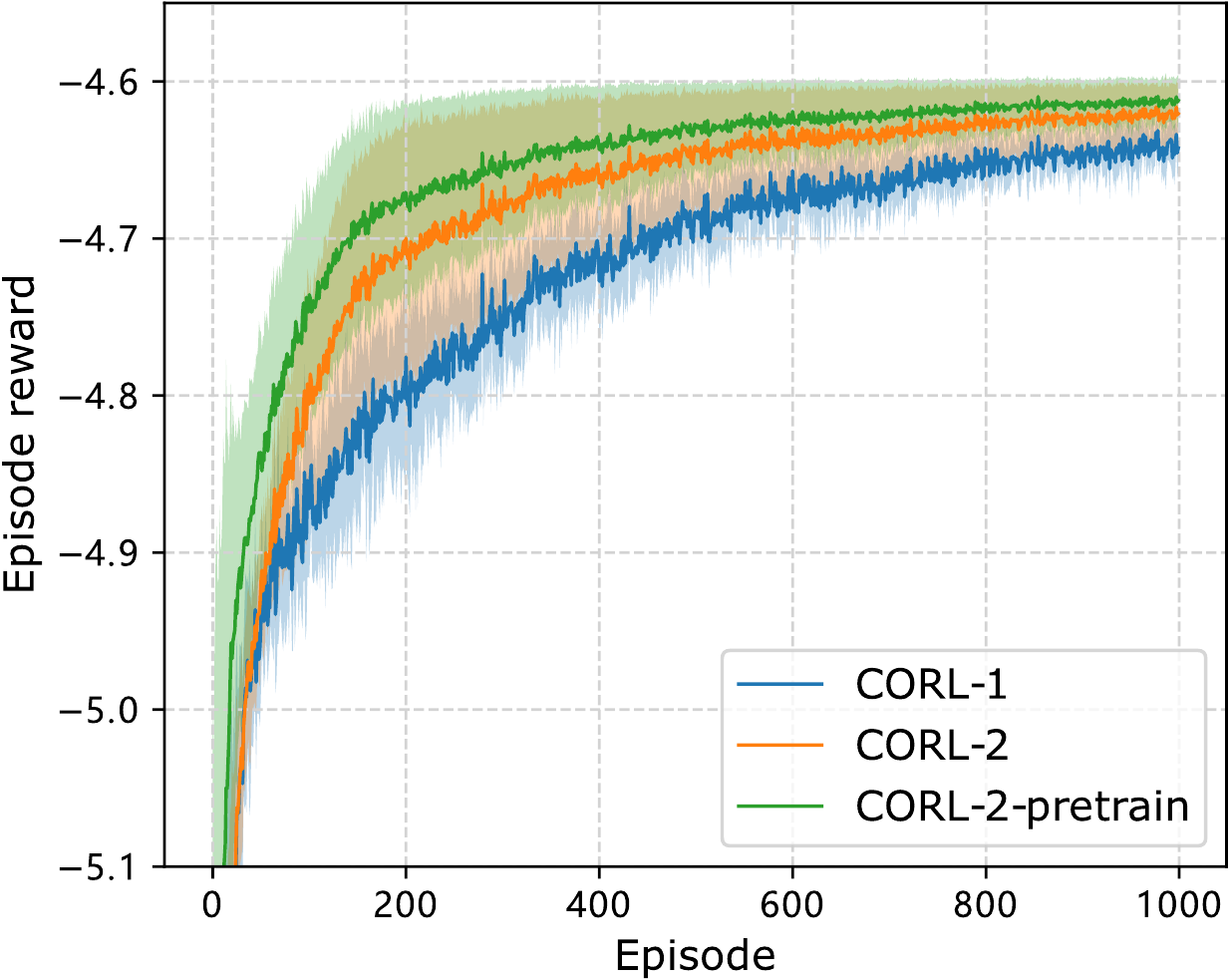}
		\caption{Learning curves of CORL-1, CORL-2 and CORL-2-pretrain on $100$-node LG data sets.}
		\label{reward-curve-corl2} 
	\end{figure}

	\begin{figure*}[ht]
		\centering
		\subfloat[{TPR on GP10}]{
			\includegraphics[width=0.35\linewidth,height=0.33\linewidth]{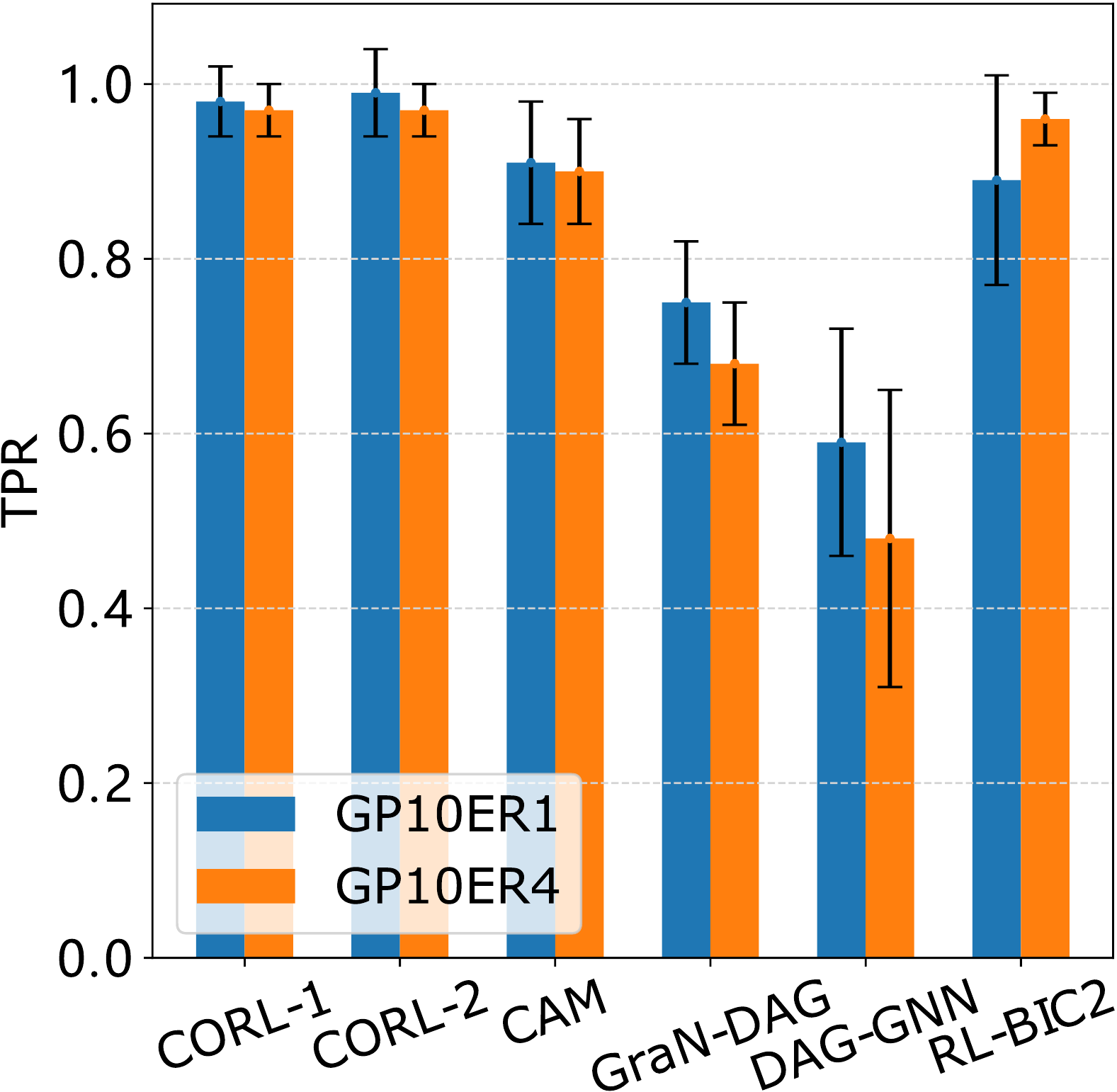}
		}\quad \quad \quad
		\subfloat[SHD on GP10]{
			\includegraphics[width=0.35\linewidth,height=0.33\linewidth]{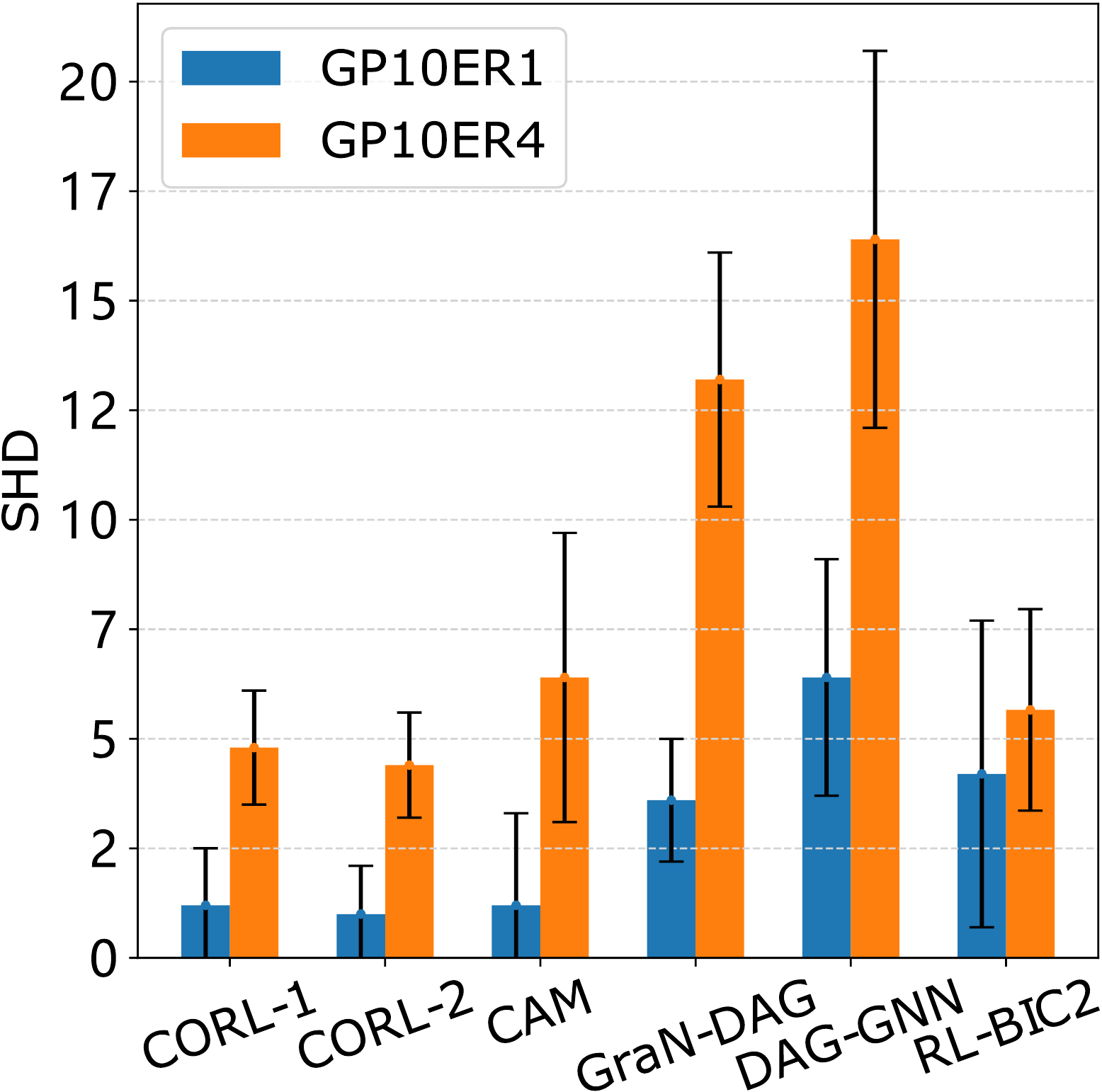}
		}\\
		\subfloat[TPR on GP30]{
			\includegraphics[width=0.35\linewidth,height=0.33\linewidth]{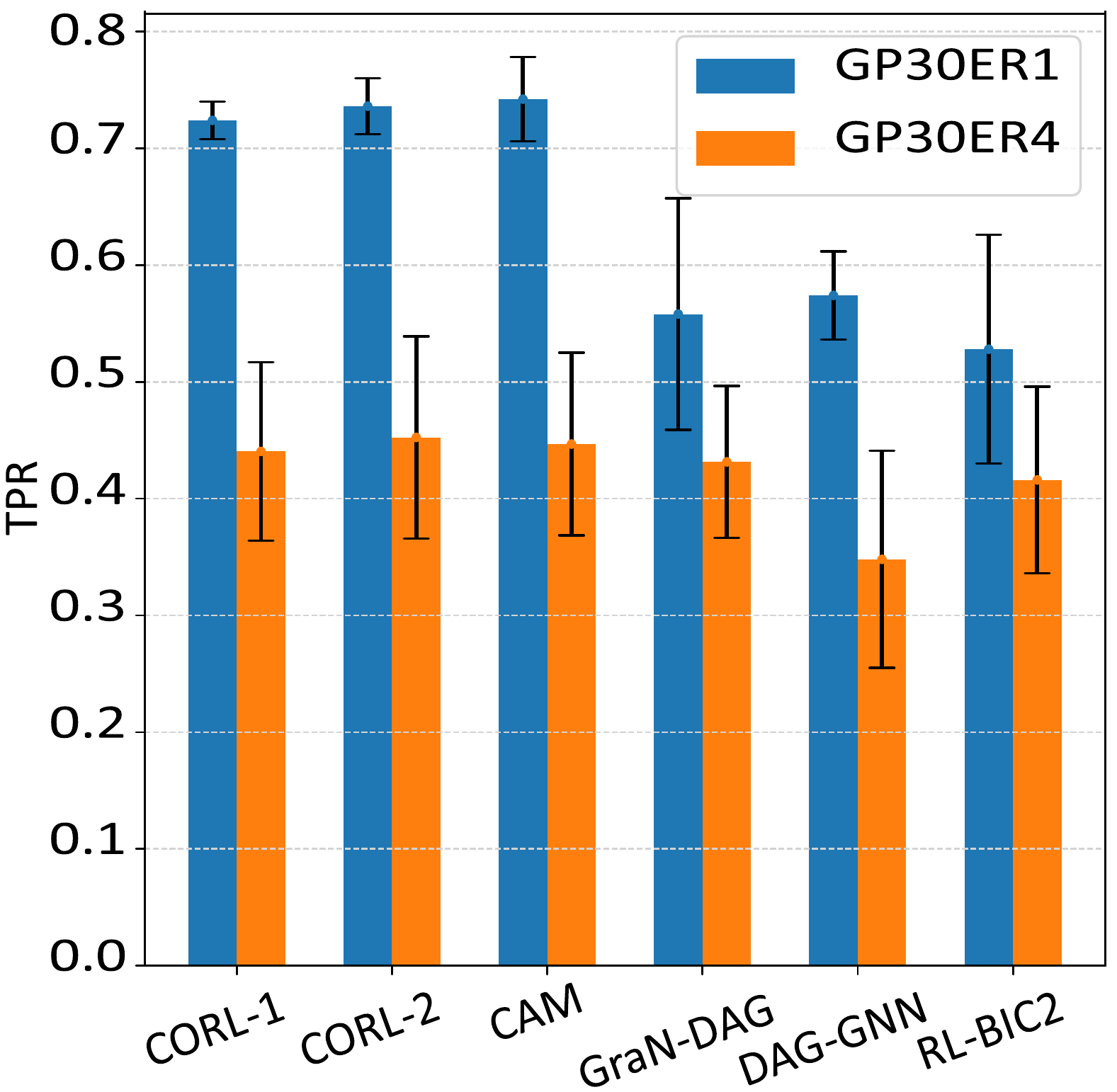}
		}\quad \quad \quad
		\subfloat[SHD on GP30]{
			\includegraphics[width=0.35\linewidth,height=0.33\linewidth]{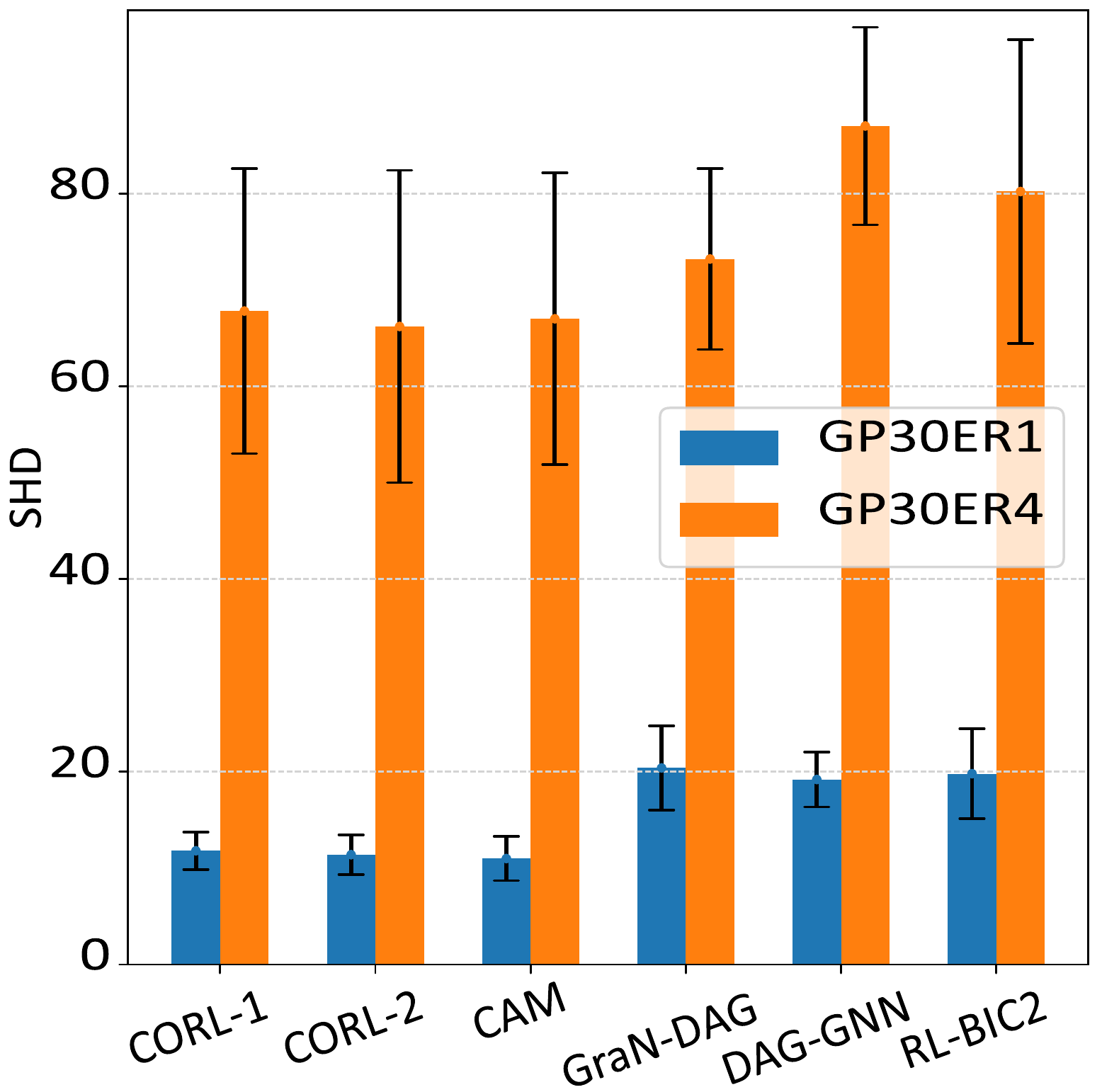}
		}
		\caption{The empirical results on GP data models with 10 and 30 nodes.} 
		\label{nonlinear_10_30}
	\end{figure*}
	
	In this section, we conduct experiments on synthetic data sets with linear and non-linear causal relationships as well as a real data set. 
	The baselines are {ICA-LiNGAM} \cite{shimizu2006linear}, three ordering-based approaches L1OBS \cite{schmidt2007learning}, CAM \cite{buhlmann2014cam} and A* Lasso \cite{Xiang2013lasso}, some recent gradient-based approaches NOTEARS \cite{zheng2018dags}, {DAG-GNN} \cite{yu19dag} and  {GraN-DAG} \cite{Lachapelle2019grandag}, and the RL-based approach {RL-BIC2} \cite{zhu2020causal}. 
	We use the original implementations (see Appendix B.1 for details) and pick the recommended hyper-parameters unless otherwise stated. 
	
	We generate different types of synthetic data sets which vary along: level of edge sparsity, graph type, number of nodes, causal functions and sample size. Two types of graph sampling schemes, Erdös–Rényi (ER) and Scale-free (SF), are considered here. 
	We denote $d$-node ER and SF graphs with on average $hd$ edges as ER$h$ and SF$h$, respectively. 
	Two common metrics are considered: True Positive Rate ({TPR}) and Structural Hamming Distance ({SHD}).   The former indicates the probability of correctly finding the positive edges among the discoveries. Hence, it can be used to measure the quality of an ordering, and the higher the better. The latter counts the total number of missing, falsely detected or reversed edges, and the smaller the better.
	
	
	
	\subsection{Linear Models with Gaussian and Non-Gaussian Noise}
	
	We evaluate the proposed methods on Linear Gaussian (LG) with equal variance Gaussian noise and LiNGAM data models, and the true DAGs in both
	cases are known to be identifiable \cite{peters2014identifiability,shimizu2006linear}. 
	We set $h \in \{2, 5\}$ and $d \in \{30,50, 100\}$  to generate  observed data (see Appendix~B.2 for details).  
	For variable selection, we set the thresholding as $0.3$ and apply it to the estimated coefficients, as similarly used by \cite{zheng2018dags,zhu2020causal}.

	Table~\ref{30-Lin_results} {presents} the  results for $30$- and $100$-node LG data models; the conclusions do not change with $50$-node  graphs, which 
	are given in Appendix~D. 
	The performances of ICA-LiNGAM, GraN-DAG and CAM is also given in Appendix~D,  and they are almost never on par with the best methods presented in this section. CORL-1 and CORL-2 achieve consistently good results on LiNGAM data sets which are reported in Appendix~E due to the space limit. 
	
	We now examine Table~\ref{30-Lin_results} (the values in parentheses represent the standard deviation across data sets per task). Across all settings, CORL-1 and CORL-2 are the best performing methods in terms of both TPR and SHD, while NOTEARS and DAG-GNN are not too far behind.  In Figure~\ref{reward-curve-corl2}, we further show the training reward curves of CORL-1 and CORL-2 on $100$-node LG data sets, where CORL-2 converges faster to a better ordering than CORL-1. We conjecture that this is because dense rewards can provide more guidance information for the training process than  episodic rewards, 
	which is beneficial  to the learning of RL model and improves the training performance. {Hence, CORL-2 is preferred in practice if the score function is decomposable for each variable. }
	As discussed previously, RL-BIC2 only achieves satisfactory results on graphs with $30$ nodes.
	The TPR of L1OBS is lower than that of A* Lasso, which indicates that L1OBS using greedy hill-climbing with tabu lists may not find a good ordering. 
	Note that the SHD of L1OBS and A* Lasso reported here are the results after applying the introduced pruning method. We observe that the SHDs are greatly improved after pruning. For example, the SHDs of L1OBS decrease from $171.6$ $(29.5)$, $588.0$ $(66.2)$ and $1964.5$ $(136.6)$ to $85.2$ $(23.8)$, $215.4$ $(26.3)$ and $481.2$ $(49.9)$ for ER2 graphs with $30$, $50$ and $100$ nodes, respectively, while the TPRs almost keep the same.
	We have also evaluated our method on $150$-node LG  data models on ER2 graphs. CORL-1 has TPR and SHD being  $0.95$ $(0.01)$ and $63.7$ $(9.1)$, while CORL-2 has $0.97$ $(0.01)$ and $38.3$ $(14.3)$, respectively. CORL-2  outperforms NOTEARS that achieves $0.94$ $(0.02)$ and $50.8$ $(21.8)$. 

	\paragraph{Pretraining.} We show the training reward curve of CORL-2-pretrain in Figure~\ref{reward-curve-corl2}, where the model parameters are pretrained in a supervised manner. The data sets used for pretraining contain $30$-node ER2 and SF2 graphs with different causal relationships. Note that the data sets used for evaluation are different from those used for pretraining. Compared to that of CORL-2 using random initialization, a pretrained model can accelerate the model  learning process.  Although pretraining requires additional time, it is only
	carried out once and when finished, the pretrained model can be used for multiple causal discovery tasks.  
	Similar conclusion can be drawn in terms of CORL-1, which is shown in Appendix~G. 
	
	\paragraph{Running time.} { We also report the running time of all the methods on $30$- and $100$-node linear data models: CORL-1, CORL-2, GraN-DAG and DAG-GNN $\approx$ $15$ minutes for $30$-node graphs; CORL-1 and CORL-2 $\approx$ $7$ hours against GraN-DAG and DAG-GNN $\approx$ $4$ hours for $100$-node graphs;
		CAM $\approx$ $15$ minutes for both $30$- and $100$-node graphs, while L1OBS and A* Lasso $\approx$ $2$ minutes for that tasks; 
		NOTEARS $\approx$ $5$ minutes and $\approx$ $1$ hour for the two tasks 
		respectively; RL-BIC2 $\approx$ $3$ hours for $30$-node graphs. 
		We set the maximal running time up to $15$ hours, but RL-BIC2 did not converge on $100$-node graphs, hence we did not report its results. Note that the running time can be significantly reduced  by paralleling the evaluation of reward.
		The neural network based learning methods 
		generally take longer time,  
		and the proposed method achieves the best performance among these methods.
	}
	
	
	
	\subsection{Non-Linear Model with Gaussian Process}

	In this experiment, we 
	consider causal relationships with $f_j$ being a function sampled from a Gaussian Process (GP) with radial basis function  kernel of bandwidth one. The additive noise follows standard Gaussian distribution, which is known to be identifiable  \cite{10.5555/2627435.2670315}. 
	We consider ER1 and ER4 graphs with different sample numbers (see Appendix~B.2 for the generation of data sets), and  we only report the results with $m=500$ samples due to the space limit (the remaining results are given in Appendix~F). 
	For comparison, only the methods that have been shown competitive for this non-linear data model in existing works \cite{zhu2020causal,Lachapelle2019grandag} are included. For a given ordering, we follow \cite{zhu2020causal} to use GP regression to fit the causal relationships. 
	{We also set a maximum time limit of $15$ hours for all the methods for fair comparison and only  graphs with up to $30$ nodes are considered here}, as using GP regression to calculate the scores is time-consuming. 
	The variable selection method used here is the CAM pruning from \cite{buhlmann2014cam}.
	
	The results on $10$- and $30$-node data sets with ER1 and ER4 graphs are shown in Figure~\ref{nonlinear_10_30}. 
	Overall, both GraN-DAG and DAG-GNN perform worse than CAM. We conjecture that this is because the  number of samples are not sufficient for GraN-DAG
	and DAG-GNN to fit neural networks well, as also shown by \cite{Lachapelle2019grandag}. CAM, CORL-1, and CORL-2 have similar results, with CORL-2 performing the best on $10$-node graphs and being slightly worse than CAM on $30$-node graphs. All of these methods have better results on ER1 graphs than on ER4 graphs, especially with $30$ nodes. We also notice that CORL-2 only runs about $700$ iterations on $30$-node graphs and about $5000$ iterations on $10$-node graphs within the time limit, due to the increased time from GP regression. Nonetheless, the proposed method achieves a much improved performance  compared with the existing RL-based method.

	\subsection{Real Data}
	The Sachs data set \cite{sachs2005causal}, with $11$-node and $17$-edge true graph,  is widely used for research on graphical models. The expression levels of protein and phospholipid in the data set can be used to discover the implicit protein signal network. 
	The observational data set has $m = 853$ samples and is used to discover the causal structure. 
	We similary use Gaussian Process regression to model the causal relationships in calculating the score.
	In this experiment,
	CORL-1, CORL-2 and RL-BIC2 achieve the best SHD $11$. 
	CAM, GraN-DAG, and  ICA-LiNGAM achieve SHDs $12$, $13$ and $14$, respectively.
	Particularly, DAG-GNN and NOTEARS result in SHDs $16$ and $19$, respectively, whereas an empty graph has SHD $17$.

	\section{Conclusion}
	
	In this work, we have incorporated RL into  the  ordering-based  paradigm for causal discovery, where a generated ordering can be pruned by variable selection to obtain the causal DAG. Two methods are developed based on the MDP formulation and an encoder-decoder framework. We further analyze the consistency and computational complexity for the proposed approach.  Empirical results validate the improved performance over existing RL-based causal discovery approach.
	
	\section*{Acknowledgments} 
	{Xiaoqiang Wang and Liangjun Ke were supported  by the National Natural Science Foundation of China under Grant 61973244.}
	
	{
		\bibliographystyle{named}
		\bibliography{ijcai21}}

\begin{thebibliography}{}

\bibitem[\protect\citeauthoryear{Bartlett and
  Cussens}{2017}]{bartlett2017integer}
Mark Bartlett and James Cussens.
\newblock Integer linear programming for the bayesian network structure
  learning problem.
\newblock {\em Artificial Intelligence}, 244:258--271, 2017.

\bibitem[\protect\citeauthoryear{Bello \bgroup \em et al.\egroup
  }{2016}]{bello2016neural}
Irwan Bello, Hieu Pham, Quoc~V Le, Mohammad Norouzi, and Samy Bengio.
\newblock Neural combinatorial optimization with reinforcement learning.
\newblock {\em arXiv preprint arXiv:1611.09940}, 2016.

\bibitem[\protect\citeauthoryear{Bernstein \bgroup \em et al.\egroup
  }{2020}]{bernstein2020ordering}
Daniel Bernstein, Basil Saeed, Chandler Squires, and Caroline Uhler.
\newblock Ordering-based causal structure learning in the presence of latent
  variables.
\newblock In {\em International Conference on Artificial Intelligence and
  Statistics (AISTATS)}, pages 4098--4108. PMLR, 2020.

\bibitem[\protect\citeauthoryear{B{\"u}hlmann \bgroup \em et al.\egroup
  }{2014}]{buhlmann2014cam}
Peter B{\"u}hlmann, Jonas Peters, Jan Ernest, et~al.
\newblock {CAM}: Causal additive models, high-dimensional order search and
  penalized regression.
\newblock {\em The Annals of Statistics}, 42(6):2526--2556, 2014.

\bibitem[\protect\citeauthoryear{Chickering}{1996}]{Chickering1996learning}
David~Maxwell Chickering.
\newblock Learning {Bayesian} networks is {NP}-complete.
\newblock In {\em Learning from Data: Artificial Intelligence and Statistics
  V}. Springer, 1996.

\bibitem[\protect\citeauthoryear{Chickering}{2002}]{chickering2002optimal}
David~Maxwell Chickering.
\newblock Optimal structure identification with greedy search.
\newblock {\em Journal of Machine Learning Research}, 3(Nov):507--554, 2002.

\bibitem[\protect\citeauthoryear{De~Campos and Ji}{2011}]{de2011efficient}
Cassio~P De~Campos and Qiang Ji.
\newblock Efficient structure learning of bayesian networks using constraints.
\newblock {\em The Journal of Machine Learning Research}, 12:663--689, 2011.

\bibitem[\protect\citeauthoryear{Friedman and Koller}{2003}]{friedman2003being}
Nir Friedman and Daphne Koller.
\newblock Being {B}ayesian about network structure. {A} {B}ayesian approach to
  structure discovery in {B}ayesian networks.
\newblock {\em Machine learning}, 50(1-2):95--125, 2003.

\bibitem[\protect\citeauthoryear{Khalil \bgroup \em et al.\egroup
  }{2017}]{khalil2017learning}
Elias Khalil, Hanjun Dai, Yuyu Zhang, Bistra Dilkina, and Le~Song.
\newblock Learning combinatorial optimization algorithms over graphs.
\newblock In {\em Advances in {N}eural {I}nformation {P}rocessing {S}ystems
  (NeurIPS)}, 2017.

\bibitem[\protect\citeauthoryear{Kool \bgroup \em et al.\egroup
  }{2019}]{kool2018attention}
Wouter Kool, Herke Van~Hoof, and Max Welling.
\newblock Attention, learn to solve routing problems!
\newblock In {\em International Conference on Learning Representations (ICLR)},
  2019.

\bibitem[\protect\citeauthoryear{Lachapelle \bgroup \em et al.\egroup
  }{2020}]{Lachapelle2019grandag}
S{\'e}bastien Lachapelle, Philippe Brouillard, Tristan Deleu, and Simon
  Lacoste-Julien.
\newblock Gradient-based neural {DAG} learning.
\newblock In {\em International Conference on Learning Representations (ICLR)},
  2020.

\bibitem[\protect\citeauthoryear{Larranaga \bgroup \em et al.\egroup
  }{1996}]{larranaga1996learning}
Pedro Larranaga, Cindy~MH Kuijpers, Roberto~H Murga, and Yosu Yurramendi.
\newblock Learning bayesian network structures by searching for the best
  ordering with genetic algorithms.
\newblock {\em IEEE Transactions on Systems, Man, and Cybernetics-Part A:
  Systems and Humans}, 26(4):487--493, 1996.

\bibitem[\protect\citeauthoryear{Ng \bgroup \em et al.\egroup
  }{2019a}]{Ng2019masked}
Ignavier Ng, Zhuangyan Fang, Shengyu Zhu, Zhitang Chen, and Jun Wang.
\newblock Masked gradient-based causal structure learning.
\newblock {\em arXiv preprint arXiv:1910.08527}, 2019.

\bibitem[\protect\citeauthoryear{Ng \bgroup \em et al.\egroup
  }{2019b}]{Ng2019GAE}
Ignavier Ng, Shengyu Zhu, Zhitang Chen, and Zhuangyan Fang.
\newblock A graph autoencoder approach to causal structure learning.
\newblock {\em arXiv preprint arXiv:1911.07420}, 2019.

\bibitem[\protect\citeauthoryear{Peters and
  B{\"u}hlmann}{2014}]{peters2014identifiability}
Jonas Peters and Peter B{\"u}hlmann.
\newblock Identifiability of gaussian structural equation models with equal
  error variances.
\newblock {\em Biometrika}, 101(1):219--228, 2014.

\bibitem[\protect\citeauthoryear{Peters \bgroup \em et al.\egroup
  }{2014}]{10.5555/2627435.2670315}
Jonas Peters, Joris~M. Mooij, Dominik Janzing, and Bernhard Sch\"{o}lkopf.
\newblock Causal discovery with continuous additive noise models.
\newblock {\em Journal of Machine Learning Research}, 15(1):2009–2053,
  January 2014.

\bibitem[\protect\citeauthoryear{Raskutti and
  Uhler}{2018}]{raskutti2018learning}
Garvesh Raskutti and Caroline Uhler.
\newblock Learning directed acyclic graph models based on sparsest
  permutations.
\newblock {\em Stat}, 7(1):e183, 2018.

\bibitem[\protect\citeauthoryear{Sachs \bgroup \em et al.\egroup
  }{2005}]{sachs2005causal}
Karen Sachs, Omar Perez, Dana Pe'er, Douglas~A Lauffenburger, and Garry~P
  Nolan.
\newblock Causal protein-signaling networks derived from multiparameter
  single-cell data.
\newblock {\em Science}, 308(5721):523--529, 2005.

\bibitem[\protect\citeauthoryear{Scanagatta \bgroup \em et al.\egroup
  }{2015}]{scanagatta2015learning}
Mauro Scanagatta, Cassio~P de~Campos, Giorgio Corani, and Marco Zaffalon.
\newblock Learning bayesian networks with thousands of variables.
\newblock In {\em NeurIPS}, 2015.

\bibitem[\protect\citeauthoryear{Schmidt \bgroup \em et al.\egroup
  }{2007}]{schmidt2007learning}
Mark Schmidt, Alexandru Niculescu-Mizil, Kevin Murphy, et~al.
\newblock Learning graphical model structure using {L}1-regularization paths.
\newblock In {\em Proceedings of the AAAI Conference on Artificial Intelligence
  (AAAI)}, 2007.

\bibitem[\protect\citeauthoryear{Shimizu \bgroup \em et al.\egroup
  }{2006}]{shimizu2006linear}
Shohei Shimizu, Patrik~O Hoyer, Aapo Hyv{\"a}rinen, and Antti Kerminen.
\newblock A linear non-{G}aussian acyclic model for causal discovery.
\newblock {\em Journal of Machine Learning Research}, 7(Oct):2003--2030, 2006.

\bibitem[\protect\citeauthoryear{Solus \bgroup \em et al.\egroup
  }{2017}]{solus2017consistency}
Liam Solus, Yuhao Wang, and Caroline Uhler.
\newblock Consistency guarantees for greedy permutation-based causal inference
  algorithms.
\newblock {\em arXiv preprint arXiv:1702.03530}, 2017.

\bibitem[\protect\citeauthoryear{Sutton and
  Barto}{2018}]{sutton2018reinforcement}
Richard~S Sutton and Andrew~G Barto.
\newblock {\em Reinforcement {L}earning: An {I}ntroduction}.
\newblock MIT press, 2018.

\bibitem[\protect\citeauthoryear{Teyssier and
  Koller}{2005}]{teyssier2012ordering}
Marc Teyssier and Daphne Koller.
\newblock Ordering-based search: A simple and effective algorithm for learning
  bayesian networks.
\newblock In {\em Conference on Uncertainty in Artificial Intelligence (UAI)},
  2005.

\bibitem[\protect\citeauthoryear{Vaswani \bgroup \em et al.\egroup
  }{2017}]{vaswani2017attention}
Ashish Vaswani, Noam Shazeer, Niki Parmar, Jakob Uszkoreit, Llion Jones,
  Aidan~N Gomez, {\L}ukasz Kaiser, and Illia Polosukhin.
\newblock Attention is all you need.
\newblock In {\em Advances in {N}eural {I}nformation {P}rocessing {S}ystems
  (NeurIPS)}, 2017.

\bibitem[\protect\citeauthoryear{Vinyals \bgroup \em et al.\egroup
  }{2015}]{vinyals2015pointer}
Oriol Vinyals, Meire Fortunato, and Navdeep Jaitly.
\newblock Pointer networks.
\newblock In {\em Advances in {N}eural {I}nformation {P}rocessing {S}ystems
  (NeurIPS)}, 2015.

\bibitem[\protect\citeauthoryear{Xiang and Kim}{2013}]{Xiang2013lasso}
Jing Xiang and Seyoung Kim.
\newblock A* lasso for learning a sparse bayesian network structure for
  continuous variables.
\newblock In {\em Advances in {N}eural {I}nformation {P}rocessing {S}ystems
  (NeurIPS)}, 2013.

\bibitem[\protect\citeauthoryear{Yu \bgroup \em et al.\egroup }{2019}]{yu19dag}
Yue Yu, Jie Chen, Tian Gao, and Mo~Yu.
\newblock {DAG-GNN}: {DAG} structure learning with graph neural networks.
\newblock In {\em International Conference on Machine Learning (ICML)}, 2019.

\bibitem[\protect\citeauthoryear{Zheng \bgroup \em et al.\egroup
  }{2018}]{zheng2018dags}
Xun Zheng, Bryon Aragam, Pradeep~K Ravikumar, and Eric~P Xing.
\newblock {DAGs with NO TEARS}: Continuous optimization for structure learning.
\newblock In {\em Advances in {N}eural {I}nformation {P}rocessing {S}ystems
  (NeurIPS)}, 2018.

\bibitem[\protect\citeauthoryear{Zheng \bgroup \em et al.\egroup
  }{2020}]{Zheng2019learning}
Xun Zheng, Chen Dan, Bryon Aragam, Pradeep Ravikumar, and Eric~P Xing.
\newblock Learning sparse nonparametric {DAGs}.
\newblock In {\em International Conference on Artificial Intelligence and
  Statistics (AISTATS)}, 2020.

\bibitem[\protect\citeauthoryear{Zhu \bgroup \em et al.\egroup
  }{2020}]{zhu2020causal}
Shengyu Zhu, Ignavier Ng, and Zhitang Chen.
\newblock Causal discovery with reinforcement learning.
\newblock In {\em International Conference on Learning Representations (ICLR)},
  2020.

\bibitem[\protect\citeauthoryear{Zoph and Le}{2017}]{zoph2016neural}
Barret Zoph and Quoc~V Le.
\newblock Neural architecture search with reinforcement learning.
\newblock In {\em International Conference on Learning Representations (ICLR)},
  2017.

\end{thebibliography}


\begin{thebibliography}{}

\bibitem[\protect\citeauthoryear{Peters and
  B{\"u}hlmann}{2014}]{peters2014identifiability}
Jonas Peters and Peter B{\"u}hlmann.
\newblock Identifiability of gaussian structural equation models with equal
  error variances.
\newblock {\em Biometrika}, 101(1):219--228, 2014.

\bibitem[\protect\citeauthoryear{Peters \bgroup \em et al.\egroup
  }{2014}]{10.5555/2627435.2670315}
Jonas Peters, Joris~M. Mooij, Dominik Janzing, and Bernhard Sch\"{o}lkopf.
\newblock Causal discovery with continuous additive noise models.
\newblock {\em Journal of Machine Learning Research}, 15(1):2009–2053,
  January 2014.

\bibitem[\protect\citeauthoryear{Shimizu \bgroup \em et al.\egroup
  }{2006}]{shimizu2006linear}
Shohei Shimizu, Patrik~O Hoyer, Aapo Hyv{\"a}rinen, and Antti Kerminen.
\newblock A linear non-{G}aussian acyclic model for causal discovery.
\newblock {\em Journal of Machine Learning Research}, 7(Oct):2003--2030, 2006.

\bibitem[\protect\citeauthoryear{Vinyals \bgroup \em et al.\egroup
  }{2015}]{vinyals2015pointer}
Oriol Vinyals, Meire Fortunato, and Navdeep Jaitly.
\newblock Pointer networks.
\newblock In {\em Advances in {N}eural {I}nformation {P}rocessing {S}ystems
  (NeurIPS)}, 2015.

\bibitem[\protect\citeauthoryear{Zhu \bgroup \em et al.\egroup
  }{2020}]{zhu2020causal}
Shengyu Zhu, Ignavier Ng, and Zhitang Chen.
\newblock Causal discovery with reinforcement learning.
\newblock In {\em International Conference on Learning Representations (ICLR)},
  2020.

\end{thebibliography}
	
\end{document}


\maketitle
\appendix
\section{Architectures and  Hyper-Parameters}

\subsection{Encoder and Decoder Architectures}  \label{multi-encoder-decoder}
There are a variety of neural network modules that can be used for encoder and decoder architectures. Here we consider some representative modules, including:  Multi Layer Perceptrons (MLP) module, an LSTM based recurrent neural network module, and the self-attention based encoder from the Transformer structure. In addition, we use the original observational data as the state directly, i.e., no encoder module is used, which is denoted as Null. More details regarding the architectures and associated hyper-parameters choices will be presented in Appendix~\ref{transformer-lstm}.


Table~\ref{encoder-decoder-results} reports the empirical results of CORL-2 on $30$-node LG ER2 data sets where the noise variances are equal (see Appendix~B.2 for details about data generation). We observe that the LSTM based decoder achieves a better performance than that of MLP based decoder, which indicates that LSTM is more effective than MLP in sequential decision tasks. The overall performance of neural network encoders is better than that of Null, which shows that the data pre-processed by an encoder module is necessary. Among all these encoders, Transformer encoder achieves the best results. Similar conclusion was drawn  in \cite{zhu2020causal}, and we hypothesize that the performance of Transformer encoder benefits from  the self-attention scheme that provide sufficient interactions amongst variables. 

\begin{table*}[ht]
  \caption{Empirical results of CORL-2 with different encoder and decoder architectures on $30$-node LG ER2 data sets. {True Positive Rate ({TPR}) indicates the probability of correctly finding the positive edges among the discoveries, and the higher the better.  Structural Hamming Distance ({SHD})  counts the total number of missing, falsely detected or reversed edges, and the smaller the better.}}
  \label{encoder-decoder-results}
  \centering \footnotesize
  \setlength{\tabcolsep}{2mm}{
  \begin{tabular}{lccccc}
  \toprule 
 &&&Encoder  \\
  \cmidrule(){3-6}
 & & Null &  LSTM &  MLP & Transformer  \\
  \cmidrule(){1-6}
   \multirow{2}{*}{MLP Decoder} & TPR & 0.81 (0.07) & {0.86 (0.10)} & {0.96 (0.02)} & 0.98 (0.02)  \\
   & SHD &54.2 (25.1) & 36.0 (26.7) & 11.0 (5.3) &  5.0 (3.3)\\
  \cmidrule(){1-6}
  \multirow{2}{*}{LSTM Decoder}
    & TPR & 0.94 (0.04) &  {0.88 (0.09)}  & 0.97 (0.01) & {0.99 (0.01)}  \\
  & SHD & 20.6 (20.0)  & 29.0 (17.8)& 8.6 (4.3) & 4.4 (3.5) \\
    \bottomrule
  \end{tabular}}
\end{table*}

\subsection{Model Architectures and Hyper-Parameters}
\label{transformer-lstm}
\begin{figure}[ht]
\centering
	\includegraphics[width=.4\textwidth,height=.3\textwidth]{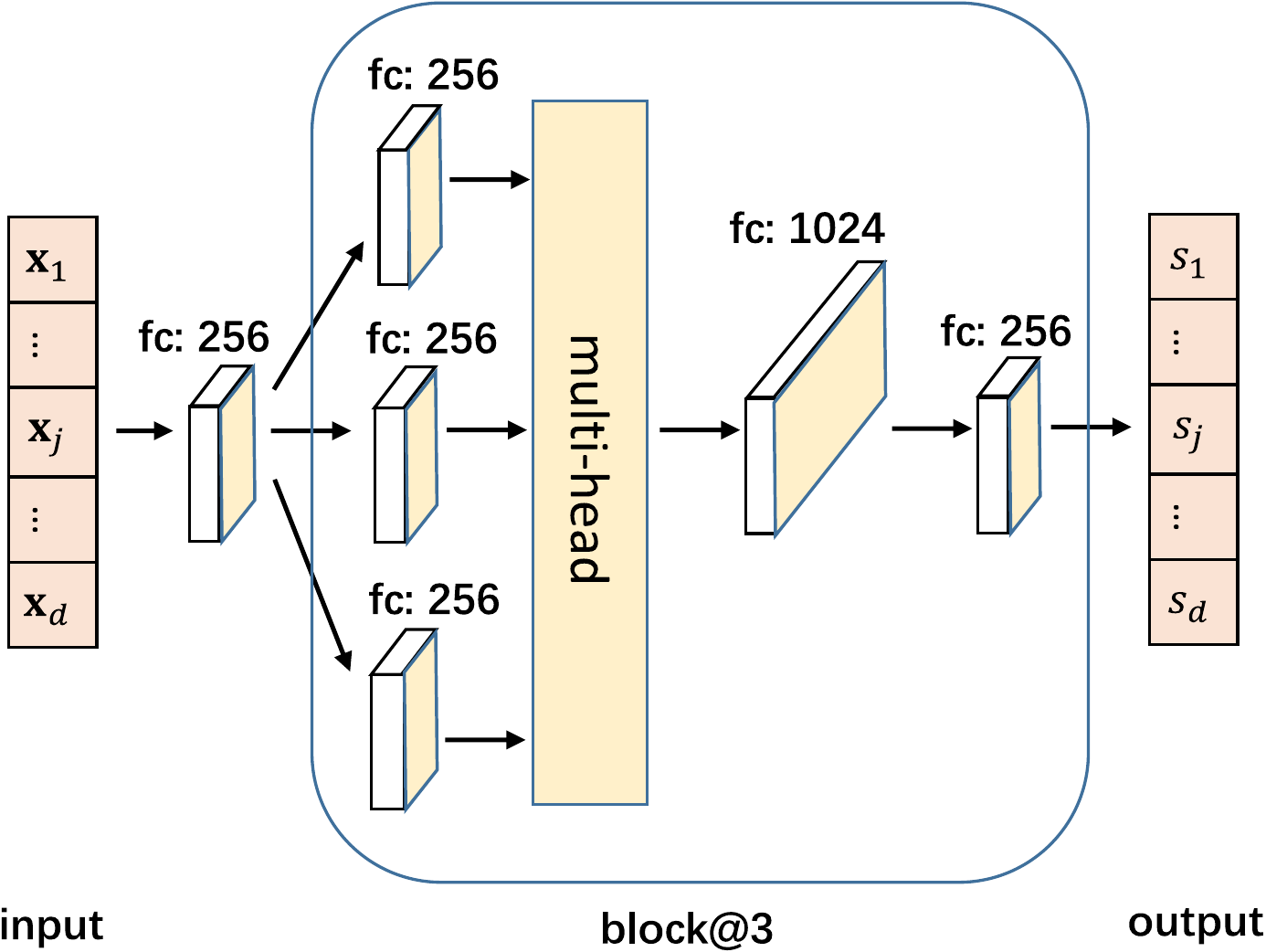}
	\caption{Illustration of the Transformer encoder. The encoder embeds the observed data $\mathrm{x}_j$ of each  variable $X_j$ into state $s_j$. Notation block@3 denotes three blocks.}
	\label{model-encoder}
\end{figure}
\begin{figure}[ht]
\centering
	\includegraphics[width=0.4\textwidth,]{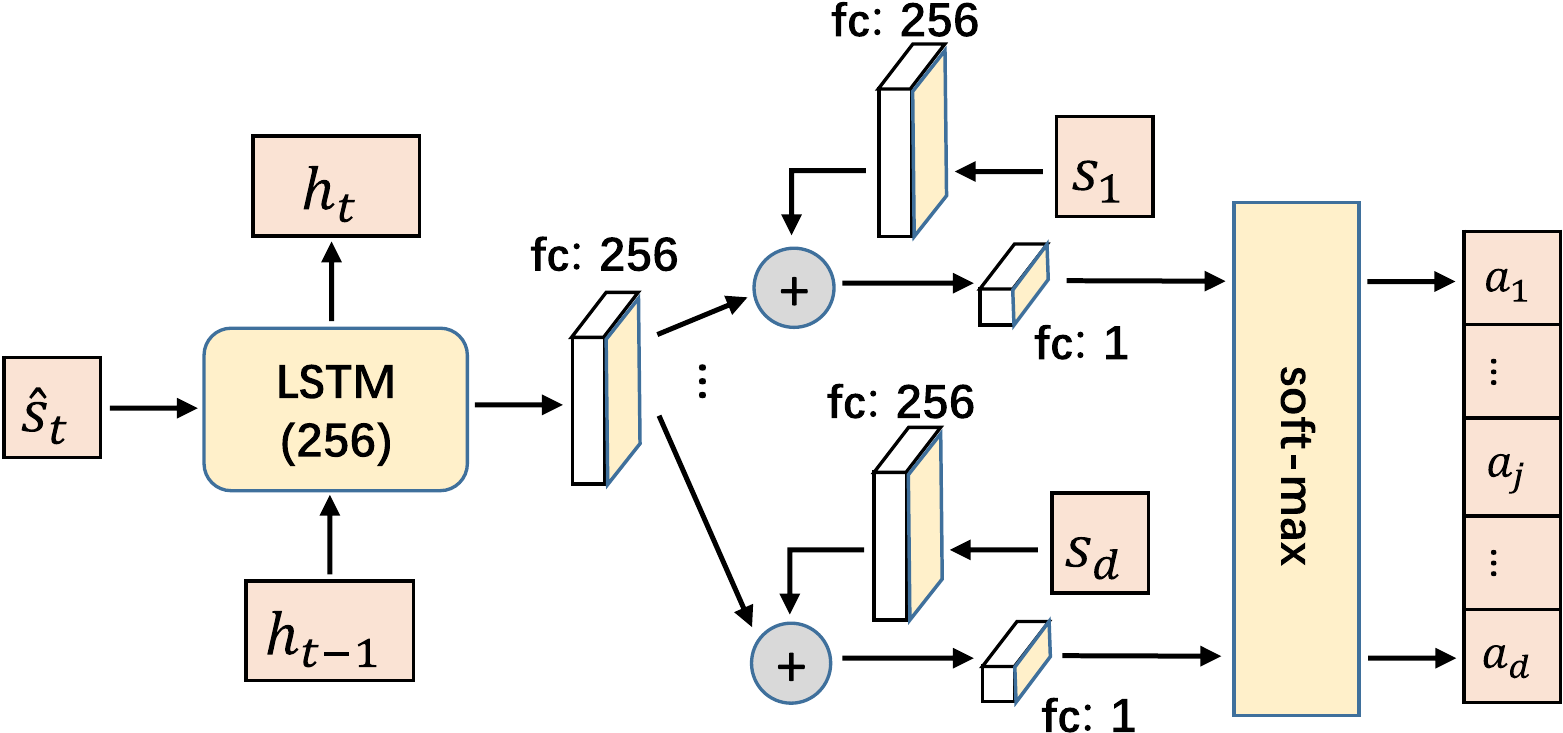}
	\caption{Illustration of LSTM decoder. At each time step, it maps the state $\hat{s}_t$ to a distribution over action space $\mathcal{A}=\{a_1,\cdots,a_d\}$ and then selects an action (variable)   according to the distribution.}
	\label{model-decoder}
\end{figure}
    
The neural network structure of the Transformer encoder  used in our experiments is given in Figure \ref{model-encoder}. It consists of a feed-forward layer with $256$ units and three blocks. Each block is composed of a multi-head attention network with $8$ heads and $2$-layer feed-forward neural networks with $1024$ and $256$ units, and each feed-forward layer is followed by a normalization layer. 
Given a batch of observed samples with  shape $b\times d \times n$, with $b$ denoting the batch size, $d$ the node number and $n$ the number of observed data in a batch, the final output of the encoder is a batch of embedded state with shape $b\times d \times 256$.

We illustrate the neural network structure of the LSTM based decoder in Figure \ref{model-decoder}, which is  similar to the decoder proposed by \cite{vinyals2015pointer}. The LSTM takes a state as input and outputs an embedding. The  embedding is mapped to the action space $\mathcal{A}$ by using some feed-forward neural networks, a soft-max module and the pointer mechanism \cite{vinyals2015pointer}. The LSTM module with $256$ hidden units is used here. 
The outputs of encoder are processed  as the initial hidden state $h_0$ for the decoder. All of the  feed-forward neural networks used in the decoder have $256$ units. For LSTM encoder, only the standard LSTM module with $256$ hidden units is used and its output is treated as the embedded state.

The MLP module consists of $3$-layer feed-forward neural networks with $256$, $512$ and $256$ units. For MLP encoder, only the standard MLP module is used. For MLP decoder, its structure is similar to Figure~2, in which LSTM module is replaced by the MLP module.

Both CORL-1 and CORL-2 use the actor-critic algorithm to train the model parameters. We use the Adam optimizer with learning rate $1e{-4}$ and $1e{-3}$ for the actor and critic, respectively.
The discount factor $\gamma$ is set to $0.98$. The actor consists of an encoder and a decoder whose choices have been described above.
The critic uses $3$-layer feed-forward neural networks with $512$, $256$, and $1$ units, which takes a state $\hat{s}$ as input and outputs a predicted value for the current policy given state $\hat{s}$.
For CORL-1, the critic needs to predict the score for each state $\hat{s}_t$, while for CORL-2, the critic takes the initial state $\hat{s}_0$ as input and outputs a predicted value directly for a complete ordering.

\section{Baselines and Date Sets}
\subsection{Baselines}\label{appendix-baselines}

The baselines considered in our experiments are listed as follows:
\begin{itemize}
    \item ICA-LiNGAM assumes linear non-Gaussian additive model for data generating procedure and applies independent component analysis to recover the weighted adjacency matrix. This method usually achieves good  performance on LiNGAM data sets. However, it does not provide guarantee for linear Gaussian data sets.\footnote{\url{https://sites.google.com/site/sshimizu06/lingam}}
    
    \item NOTEARS recovers the causal graph by estimating the weighted adjacency matrix with the least squares loss and a smooth characterization for acyclicity constraint.\footnote{
    \url{https://github.com/xunzheng/notears }}
    \item DAG-GNN formulates causal discovery in the framework of variational autoencoder. It uses a modified smooth characterization for acyclicity and optimizes a weighted adjacency matrix with the evidence lower bound as loss function.\footnote{\url{https://github.com/fishmoon1234/DAG-GNN}} 

    \item GraN-DAG models the conditional distribution of each variable given its parents with feed-forward neural networks. It also uses the smooth acyclicity constraint from NOTEARS to find a DAG that maximizes the log-likelihood of the observed samples.\footnote{\url{https://github.com/kurowasan/GraN-DAG}} 

    \item RL-BIC2 formulates the causal discovery as a one-step decision making process, and combines the score function and acyclicity constraint from NOTEARS as the reward for a directed graph.\footnote{\url{https://github.com/huawei-noah/trustworthyAI/tree/master/Causal_Structure_Learning/Causal_Discovery_RL}}
\item CAM conducts a greedy estimation procedure that starts with an empty
DAG and adds at each iteration the edge $(v_k, v_j)$ between nodes $v_k$ and $v_j$ that corresponds to the largest gain in log-likelihood. For a searched ordering, CAM prunes it to the final DAG by applying significance testing of covariates. CAM also performs  preliminary neighborhood selection to reduce the ordering space.\footnote{\url{https://cran.r-project.org/web/packages/CAM.}}
    
    \item L1OBS performs heuristic search (greedy hill-climbing with tabu lists) through the space of topological orderings to see an ordering with the best score. It uses $\ell_1$ variable selection to prune the searched ordering (fully-connected DAG) to the final DAG.\footnote{\url{https://www.cs.ubc.ca/~murphyk/Software/DAGlearn/}}
    
    \item A* Lasso with a limited queue size incorporates a heuristic scheme into a dynamic programming based method. The queue size usually needs to be adjusted to balance the time cost and the quality of the solution.\footnote{\url{http://www.cs.cmu.edu/~jingx/software/AstarLasso.zip}}
    \end{itemize}

\subsection{Data Generation} \label{data-gener}


We generate synthetic data sets which vary along five dimensions: level of edge sparsity, graph type, number of nodes, causal functions and sample size.
We sample $5$ data sets with a required number of samples for each task: a ground truth DAG $\mathcal{G}^*$ is firstly drawn randomly from either the Erdös–Rényi (ER) or Scale-free (SF) graph model, and the data are then generated according to different given SEMs. 

Specifically, for Linear Gaussian (LG) models, 
we set $h \in \{2, 5\}$ and $d \in \{30,50, 100\}$ to obtain the ER  and SF graphs  with different levels of edge sparsity and different numbers of nodes. Then we generate $3,000$ samples for each task following the linear SEM: $\mathbf{X} = W^T \mathbf{X} + \epsilon$, where $W\in \mathbb{R}^{d\times d}$ denotes the weighted adjacency matrix obtained by assigning edge weights independently from $\text{Unif}([-2, -0.5]\cup[0.5, 2])$. Here $\epsilon\in\mathbb R^d$ denote standard Gaussian noises with equal variances for each variable, which makes $\mathcal G^*$ identifiable \cite{peters2014identifiability}. 

For LiNGAM data model, the  data sets are generated similarly to LG data models but  the noise variables follow  non-Gaussian distributions  which pass the noise samples from Gaussian distribution through a power nonlinearity to make them non-Gaussian \cite{shimizu2006linear}. LiNGAM is also identifiable, as shown in \cite{shimizu2006linear}.

The GP data sets with different sample sizes are generated following $X_j = f_j(\text{Pa}(X_j)) + \epsilon_j$,   where the function $f_j$ is a function sampled from a GP with radial basis function  kernel of bandwidth one and $\epsilon_j$ follows standard Gaussian distribution.
This setting is also identifiable according to \cite{10.5555/2627435.2670315}.
Due to the efficiency of the reward calculation, we only conduct experiments with up to $30$ variables.


\section{Number of Episodes Before Convergence}\label{section-iteration}

Table~\ref{iteration-number} reports the total number of episodes required for CORL-2 and RL-BIC2 to converge, averaged over five
seeds. CORL-2 performs fewer episodes than RL-BIC2 to reach a convergence, which we believe is  due to reducing the size of search space and avoiding dealing with acyclicity.
  
\begin{table*}[ht]
  \caption{Total number of iterations ($\times10^3$) to reach convergence.}
  \label{iteration-number}
  \centering \footnotesize
  \setlength{\tabcolsep}{2mm}{
  \begin{tabular}{lcccccc}
  \toprule 
 & \multicolumn{2}{c}{30 nodes} & \multicolumn{2}{c}{50 nodes } & \multicolumn{2}{c}{100 nodes } \\
  \cmidrule(){2-7}
  & ER2 & ER5 &  ER2 &   ER5 &  ER2 &  ER5  \\
  \cmidrule(){1-7}
 CORL-2 & 1.0 (0.3)  & 1.1 (0.4) & 1.9 (0.3) & {2.4 (0.3)} & {2.3 (0.5)} & 2.9 (0.4)  \\
 RL-BIC2 &  3.9 (0.5) & 4.1 (0.6) & 3.4 (0.4) &  {3.5 (0.5)}  & $\times$ &  $\times$  \\
    \bottomrule
  \end{tabular}}
\end{table*}

\section{Additional Results  on LG Data Sets}
\label{Linear-additional}
The results for $50$-node LG data models are presented in Table~\ref{50-Lin_results}. The conclusion is similar to that of the $30$- and
$100$-node experiments. 
The results of ICA-LiNGAM, GraN-DAG and CAM  on LG data models are presented in  Table~\ref{otherBaselines-Lin_results}. Their performances do not compare favorably to CORL-1 nor CORL-2 on LG data sets.
It is not surprising that ICA-LiNGAM does not perform well because the algorithm is specifically designed for non-Gaussian noise and does not provide guarantee for LG  data models. 
We hypothesize that CAM's poor performance on LG data models is because it uses nonlinear regression instead of linear regression. As for GraN-DAG, it uses $2$-layer feed-forward neural networks to model the causal relationships, which may not be able to learn a good linear relationship in this experiment. 


\begin{table*}[ht]
  \caption{Empirical results for ER and SF graphs of $50$ nodes with LG data. The higher TPR the better, the smaller SHD the better.}
  \label{50-Lin_results}
  \centering \footnotesize
  \begin{tabular}{lccccccccc}
  \toprule 
    & & RANDOM & NOTEARS & DAG-GNN & RL-BIC2  & L1OBS & A{*} Lasso & CORL-1 & CORL-2 \\
    \midrule
  \multirow{2}{*}{ER2} & TPR & 0.31 (0.03) & 0.94 (0.02) & 0.94 (0.04) & 0.79 (0.10) &  0.56 (0.02) & 0.88 (0.03) & \bf{0.97 (0.04)} & \bf{0.97 (0.02)} \\
    & SHD & 295.4 (28.5) & 38.6 (10.8) & 30.6 (8.3) & 88.5 (49.3) & 288.0 (66.2) &154.3 (27.6) &\bf{24.0 (32.3} & \bf{17.9 (10.6)}\\
    \midrule
  \multirow{2}{*}{ER5} & TPR& 0.32 (0.02) &\bf{0.90 (0.01)} & 0.87 (0.14)& {0.74 (0.03)} & {0.57 (0.03)} & {0.82 (0.03)} & \bf{0.90 (0.02)} & \bf{0.92 (0.02)} \\
   & SHD& 378.4 (24.2)  &  \bf{67.8 (7.5)} & 93.2 (109.4)   & {128.9 (40.4)} & 299.4 (53.6) &{104.0 (28.3)} &\bf{68.3 (10.2)} & \bf{64.8 (13.1)}\\
    \midrule
 \multirow{2}{*}{SF2} & TPR & 0.49 (0.04) & 0.99 (0.01) & 0.90 (0.13)& {0.84 (0.05)} & {0.67 (0.02)} & {0.89 (0.03)} & \bf{1.00 (0.00)} & \bf{1.00 (0.00)} \\
& SHD & 215.6 (14.7) & 3.5 (1.6) & {79.3 (93.2)} & {115.2 (57.4)}   & {182.3 (33.4)} & {124.0 (35.2)} &\bf{0.0 (0.0)} & \bf{0.0 (0.0)}\\
\midrule
 \multirow{2}{*}{SF5} & TPR & 0.51 (0.03) & \bf{0.94 (0.12)} & {0.88 (0.12)} & {0.75 (0.05)} & {0.61 (0.03)} & {0.81 (0.02)} & \bf{0.94 (0.03)} & \bf{0.95 (0.02)} \\
& SHD & 345.6 (24.3) & \bf{20.1 (14.3)} & {89.2 (99.2)} & {115.2 (57.4)}   & {217.3 (36.4)} &{131.0 (25.3)} &\bf{24.3 (11.2)} & \bf{20.8 (10.1)}\\
    \bottomrule
  \end{tabular}
\end{table*}  

\begin{table*}[ht]
  \caption{Empirical results of  ICA-LiNGAM,  GraN-DAG and CAM  (against CORL-2 for reference) for ER and SF graphs with LG data.}
  \label{otherBaselines-Lin_results}
  \centering \footnotesize
  \begin{tabular}{lcccccc}
  \toprule 
    & &  & ICA-LiNGAM & GraN-DAG & CAM  & CORL-2 \\
    \midrule
 \multirow{8}{*}{30 nodes} & \multirow{2}{*}{ER2} & TPR  & 0.75 (0.03) & 0.51 (0.17) & 0.47 (0.05) & \bf{0.99 (0.01)} \\
 &   & SHD & 112.3 (12.8) & 96.0 (11.3) & 110.8 (10.3)  & \bf{4.4 (3.5))}\\
    \cmidrule(){2-7}
  & \multirow{2}{*}{ER5} & TPR &0.57 (0.03) & {0.52 (0.03)} &  {0.46 (0.02)} & \bf{0.95 (0.03)} \\
  & & SHD & {161.8 (9.2)} & {175.2 (27.4)} & {191.3 (32.5)} & \bf{37.6 (14.5)}\\
  \cmidrule(){2-7}
 & \multirow{2}{*}{SF2} & TPR &{0.58 (0.05)} & {0.61 (0.04)} & {0.63 (0.02)} & \bf{1.0 (0.0)} \\
&& SHD & 149.0 (19,8) & {136.4 (21.2)} & {115.2 (26.7)}   & \bf{0.0 (0.0)}\\
\cmidrule(){2-7}
 & \multirow{2}{*}{SF5} & TPR &{0.56 (0.04)} &  {0.58 (0.02)} & {0.60 (0.03)} & \bf{1.0 (0.0)} \\
&& SHD &{160.5 (8.9)} & 142.4 (24.3)  & {122.2 (17.4)}  & \bf{0.0 (0.0)}\\
\midrule
 \multirow{8}{*}{50 nodes} & \multirow{2}{*}{ER2} & TPR  & 0.73 (0.03) & 0.11 (0.04)& 0.55 (0.06)  & \bf{0.97 (0.02)} \\
  &  & SHD & 108.8 (11.3) & 173.0 (22.9)  & 140.8 (35.4) & \bf{17.9 (10.6)}\\
    \cmidrule(){2-7}
  & \multirow{2}{*}{ER5} & TPR & {0.57 (0.01)} & {0.64 (0.03)} & {0.61 (0.02)} & \bf{0.92 (0.02)} \\
  & & SHD & {199.8 (90.7)} & {154.2 (36.4)} & 178.3 (34.8) & \bf{64.8 (13.1)}\\
   \cmidrule(){2-7}
& \multirow{2}{*}{SF2} & TPR &{0.59 (0.04)} & {0.44 (0.05)} & {0.57 (0.02)} & \bf{1.00 (0.00)} \\
&& SHD &{208.5 (83.2)}  & 158.6 (34.5) & {131.2 (24.4)}  &  \bf{0.0 (0.0)}\\
\cmidrule(){2-7}
& \multirow{2}{*}{SF5} & TPR &{0.57 (0.01)} &  {0.49 (0.04)} & {0.53 (0.03)} & \bf{0.95 (0.02)} \\
&& SHD & 216.6 (88.4) & {243.9 (27.2)} & {235.2 (34.2)} & \bf{20.8 (10.1)} \\
\midrule 
 \multirow{8}{*}{100 nodes} & \multirow{2}{*}{ER2} & TPR  & 0.73 (0.02)  & 0.38 (0.02) & 0.43 (0.02) & \bf{0.98 (0.01)} \\
   & & SHD & 268.4 (28.5) & 191.3 (31.9) & 126.4 (27.8) & \bf{18.6 (5.7)}\\
    \cmidrule(){2-7}
 & \multirow{2}{*}{ER5} & TPR &{0.57 (0.05)} & {0.42 (0.03)} & {0.47 (0.02)} & \bf{0.94 (0.03)} \\
  & & SHD & 311.1 (63.7) & {208.2 (54.4)} & 182.3 (34.9) & \bf{164.8 (17.1)}\\
 \cmidrule(){2-7}
& \multirow{2}{*}{SF2} & TPR & {0.69 (0.03)} & {0.40 (0.03)} & {0.44 (0.02)} & \bf{1.00 (0.00)} \\
&& SHD &367.6 (67.5) & {239.9 (43.2)} & {35.2 (37.4)} & \bf{0.0 (0.0)}\\
\cmidrule(){2-7}
& \multirow{2}{*}{SF5} & TPR &{0.57 (0.05)} &  {0.39 (0.03)} & {0.48 (0.04)} & \bf{0.98 (0.01)} \\
&& SHD & 362.3 (82.8) & {219.3 (32.2)} & {125.2 (24.7)} & \bf{10.8 (6.1)}\\
    \bottomrule
  \end{tabular}
\end{table*}  

\section{Results on LiNGAM Data Sets} \label{LiNGAM-results}

\begin{table*}[ht]
  \caption{Empirical results on $30$-, $50$- and $100$-node LiNGAM ER2 data sets.}
  \label{LiNGAM-30-50-100-Table-results}
  \centering \footnotesize
  \begin{tabular}{lcccccc}
  \toprule 
  & \multicolumn{2}{c}{30 nodes ER2} & \multicolumn{2}{c}{50 nodes ER2} & \multicolumn{2}{c}{100 nodes ER2} \\
  \cmidrule(){1-7}  
    Method & TPR  & SHD & TPR & SHD & TPR  & SHD \\
    \midrule
    \bf{ICA-LiNGAM}  & \bf{1.00 (0.00)} & \bf{0.0 (0.0)} & \bf{1.00 (0.00)} & \bf{0.0 (0.0)} & \bf{1.00 (0.00)} & \bf{1.0 (0.9)} \\
    NOTEARS  & 0.94 (0.04)  & 17.2 (13.2)  & {0.95 (0.02)}  & {33.2 (16.5)} & {0.94 (0.03)} & 69.2 (23.2)  \\
    DAG-GNN  &0.94 (0.03)&19.6 (10.5)  &{0.96 (0.01)}&{24.6 (2.9)} & 0.93 (0.03) & 66.2 (19.2)\\
     GraN-DAG & 0.28 (0.09)&100.8 (14.6)&0.20 (0.01)&177.0 (25.9) &0.16 (0.04) & 312.8 (25.2) \\
  RL-BIC2  & 0.94 (0.07)   & 19.8 (23.0)   & 0.80 (0.08)   & 86.0 (51.9) & 0.13 (0.12)   & 291.3 (24.1)  \\
  CAM  & 0.60 (0.11) & 310.0 (34.0)   & 0.33 (0.07) & 178.0 (31.9) & 0.53 (0.05) & 247.2 (32.1)  \\ 
  L1OBS & 0.72 (0.04) & 85.3 (23.3) & 0.47 (0.02) & 212.6 (24.6)  & 0.41 (0.03) & 470.5 (48.1) \\
  A{*} Lasso & 0.87 (0.03) & 42.3 (16.3) & 0.88 (0.03) & 82.6 (17.6)  & 0.85 (0.04) & 102.5 (22.6)  \\
  \midrule
    \bf{CORL-1}  & \bf{0.99 (0.01)} & \bf{3.8 (6.4)}  & \bf{0.96 (0.06)} & \bf{24.6 (37.7)} &\bf{0.98 (0.01)} & \bf{20.0 (7.9)}\\
    \bf{CORL-2} & \bf{0.99 (0.01)} & \bf{3.9 (5.6)}  & \bf{0.96 (0.08)} & \bf{20.2 (11.3)} &\bf{0.99 (0.01)} & \bf{13.8 (7.2)}\\
    \bottomrule
  \end{tabular}
\end{table*}

We report the empirical results on $30$-, $50$- and $100$-node LiNGAM data sets in Table~\ref{LiNGAM-30-50-100-Table-results}.  For L1OBS, we increased the  recommended number of evaluations, from $2,500$ to $10,000$. For A* Lasso, we pick the queue size from $\{10, 500, 1000\}$, and report the best result out of these  parameter settings. 
The results of L1OBS and A* Lasso reported here are those after pruning with the same method as used by CORL-2.    
For other baselines, we pick the recommended hyper-parameters.

Among all these algorithms, ICA-LiNGAM can recover the true graph on  most of the LiNGAM data sets. This is because  ICA-LiNGAM is specifically designed for non-Gaussian noises. 
CORL-1 and CORL-2 achieve consistently good results, compared with other baselines.

\section{Results on 20-Node GP Data Sets with Different Sample Sizes}\label{diff-smaplesize}

We take the $20$-node GP data models as an example to show the performance of our method w.r.t.~different sample numbers.  The data generated based on ER4 graphs. We illustrate the empirical results in Table~\ref{gp-20-ER1-ER4-results}. 
Since previous experiments have shown that CORL-2 is slightly better than CORL-1, we only report the results of CORL-2 here.  
We also report the results of CAM on these data sets, as it is the most competitive baseline. TPR reported here is calculated based on the variable ordering (i.e., w.r.t.~its correponding fully-connected DAG). As the sample size decreases, CORL-2 tend to perform better than CAM, and we believe this is because CORL-2 benefits from the exploration ability of RL.

\begin{table*}[ht]
  \caption{Empirical results on $20$-node GP ER1 data sets with different sample sizes.}
  \label{gp-20-ER1-ER4-results}
  \centering \footnotesize
  \setlength{\tabcolsep}{2mm}{
  \begin{tabular}{lccc}
  \toprule 
Sample size & Name    & TPR  & SHD \\
    \midrule
     \multirow{2}{*}{1000} & {CAM} & {0.91 (0.03)} & {30.0 (3.7)}  \\
    &{CORL-2}  & 0.87 (0.03) & 36.5 (3.1) \\
    \midrule
    \multirow{2}{*}{500} & {CAM} & {0.86 (0.03)} & {45.0 (2.5)}  \\
    &{CORL-2} & 0.85 (0.03) & 46.3 (2.3) \\
    \midrule
    \multirow{2}{*}{400} & CAM & 0.83 (0.02) & 51.0 (2.7)  \\
    &{CORL-2} &{ 0.84 (0.03)} & {50.5 (3.0)} \\
      \midrule
    \multirow{2}{*}{200} & CAM & 0.60 (0.03) & 66.3 (1.9)  \\
    & {CORL-2} & {0.75 (0.02)}  & {63.1 (1.5)}  \\ 
    \bottomrule
  \end{tabular}}
\end{table*}

\section{CORL-1 with a Pretrained Model}\label{corl1-pretrain}

\begin{figure}[ht]
\centering
	\includegraphics[width=0.35\textwidth,]{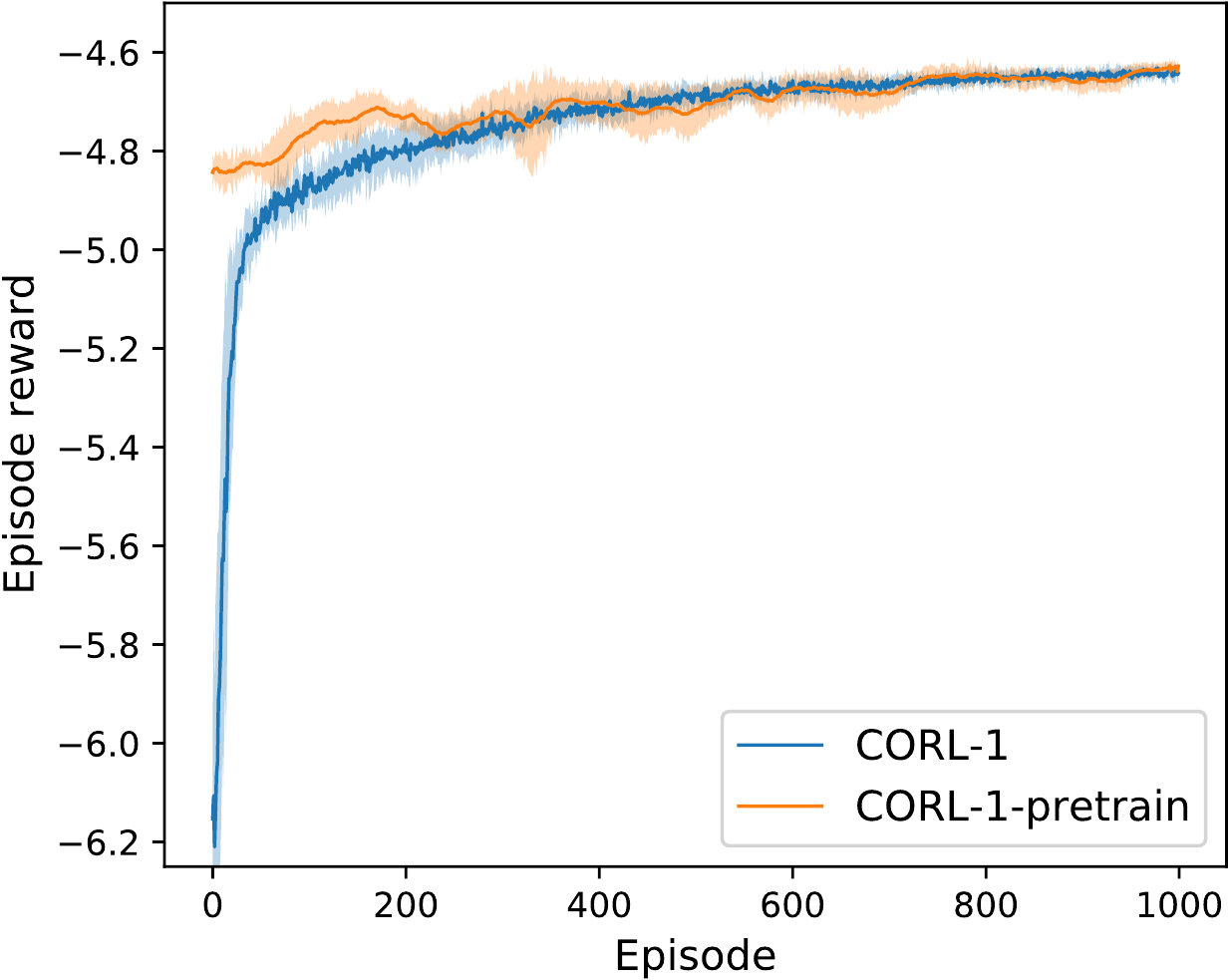}
	\caption{Learning curves of CORL-1 and CORL-1-pretrain on $100$-node LG data sets.}
	\label{corl1_pre}
\end{figure}


Figure~\ref{corl1_pre} shows the training reward curves of CORL-1 and CORL-1-pretrain; the latter stands for CORL-1 with a pretrained model. The data sets used for pretraining are $30$-node ER2 and SF2 graphs with different causal relationships.  We can observe that  the pretrained model, which serves as a good initialization, again accelerates  training.


\bibliographystyle{named}
\bibliography{ijcai21}